\documentclass{article}

\PassOptionsToPackage{marginparwidth=60pt}{geometry}

\usepackage{PRIMEarxiv}

\usepackage{wrapfig}
\usepackage[utf8]{inputenc} %
\usepackage[T1]{fontenc}    %
\usepackage{hyperref}       %
\usepackage{url}            %
\usepackage{booktabs}       %
\usepackage{amsfonts}       %
\usepackage{nicefrac}       %
\usepackage{microtype}      %
\usepackage{lipsum}
\usepackage{fancyhdr}       %
\usepackage{graphicx}       %
\graphicspath{{media/}}     %
\usepackage{subcaption}

\usepackage{amsmath, amsthm}

\usepackage [english]{babel}
\usepackage [autostyle, english = american]{csquotes}
\MakeOuterQuote{"}
\title{Don't throw the baby out with the bathwater: How and why deep learning for ARC}

\author{
  Jack Cole$^1$ $^2$, Mohamed Osman$^3$ \\
  Mindware Consulting, Inc.$^1$
  MindsAI$^2$ \\
  Tufa Labs$^3$ \\
  \texttt{ jackcole@mindware.mobi, mohamed@tufalabs.ai} \\
}

\begin{document}
\maketitle

\begin{abstract}

The Abstraction and Reasoning Corpus (ARC-AGI) presents a formidable challenge for AI systems.
Despite the typically low performance on ARC, the deep learning paradigm remains the most effective
known strategy for generating skillful (state-of-the-art) neural networks (NN) across varied
modalities and tasks in vision, language etc. The deep learning paradigm has proven to be able
to train these skillful neural networks and learn the abstractions needed in these diverse
domains.
Our work doubles down on that and continues to leverage this paradigm by incorporating on-the-fly NN training at test time.

We demonstrate that fully committing to deep learning’s capacity to acquire novel abstractions
yields state-of-the-art performance on ARC. Specifically, we treat both the neural network
and the optimizer (rather than just a pre-trained network) as integral components of the inference
process, fostering generalization to unseen tasks.

Concretely, we propose a methodology for training on ARC, starting from pretrained LLMs, and enhancing their ARC reasoning.
We also propose Test-Time Fine-Tuning (TTFT) and the Augment Inference Reverse-Augmentation and Vote (AIRV) as effective test-time techniques.
We are the first to propose and show deep learning can be used
effectively for ARC, showing boosts of up to 260\% in accuracy with AIRV and a further 300\% boost with TTFT.
An early version of this approach secured first place in the 2023 ARCathon competition, while the final version achieved the current best score on the ARC private test-set (58\%).

Our findings highlight the key ingredients of a robust reasoning system in unfamiliar domains,
underscoring the central importance mechanisms that improve broad perceptual reasoning (deep learning)—over focusing on logical inference methodologies—in
achieving high accuracy on ARC and similar domains.

\end{abstract}

\footnotetext[1]{%
The MindsAI team in the ARC-AGI 2024 competition also consisted of Michael Hodel who has had a significant impact on the team's success in 2024.
}

\section{Introduction}
Some criticisms of the current deep learning (DL) paradigm rightly note that current best models and methods are overfit
to the popular datasets \cite{Chollet2019OnTM}. The limitations of testing on large datasets have become apparent.
For example, performance on ImageNet has exceeded saturation, where models are progressing by learning patterns in the biases that labelers of ImageNet tended to make \cite{AreWeDoneWithImageNet}.
\begin{figure}

    \vspace{-25pt} %
  \centering
  \includegraphics[width=0.4\textwidth]{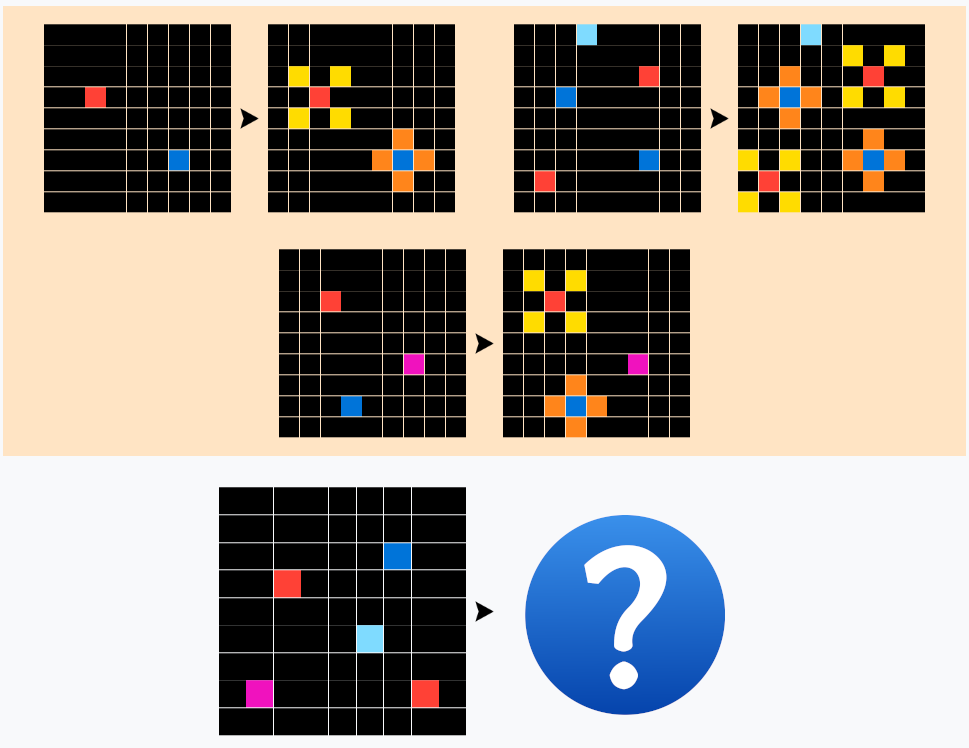}
  \caption{An example of a single easy ARC task (one datapoint).
  This task is solved by surrounding the red pixels with four yellow corners, blue pixels with four orange
  side pixels, whereas cyan and magenta input pixels remain unchanged.
}
  \label{fig:arc_example_3}

    \vspace{-15pt} %
\end{figure}
This is an indication of a gap in the current testing paradigm.
Trained models are tested on data that is similar to the data they were trained on (in-distribution data).
Models are evaluated on existing skills and we neglect benchmarking the efficiency of the learning process itself \cite{Chollet2019OnTM}.

Alternatively, consider the example of this single riddle from the Abstraction and Reasoning Corpus (ARC) \cite{Chollet2019OnTM}, shown in \autoref{fig:arc_example_3}.
Any system attempting to solve the riddle must work from the three provided examples and should infer that it is necessary to transform each colored square into its corresponding pattern.
Due to the simplicity of this dataset's setup, little pre-training knowledge can be leveraged to solve these riddles.
Instead, the solver must "figure out" the transformation based on the few examples provided.
Crucially, the model must learn a new transformation, which limits the degree to which the model can rely on pretrained knowledge or zero shot performance.
This puts a heavy emphasis on the need for contextual reasoning (reaching the correct associations) during the evaluation.
Therefore, this dataset becomes a reliable test of the efficiency of the learning process itself.

Large-scale foundation models (both vision based and LLMs) do not perform well out of the box,
when prompted with these problems \cite{concept_arc_benchmark, general_pattern_machines}.
Moreover, even though neural nets generally achieve state-of-the-art in natural
language processing and difficult visual classification and detection tasks \cite{Devlin2019BERTPO, VIT},
these are perceptual/qualitative-type problems that require highly contextual and dynamic reasoning.
There is significant uncertainty in the research community on whether neural nets can ever be
trained or enhanced to perform well on ARC or similar datasets \cite{Chollet2019OnTM, concept_arc_benchmark}.

Our perspective that these ARC problems are actually more "perceptual"
and qualitative than quantitative in nature.
There are no performant quantitative approaches to searching the space of possible input to output transformations since there are near infinite possible transformations even with only a few basic priors \cite{Chollet2019OnTM}.

Learning performant abstractions from data is what the deep learning paradigm is known for,
specifically when applied on difficult perceptual or qualitative problems.
Deep learning solutions produce state-of-the-art results on perceptual problems, for example NLP and vision.
The deep learning paradigm consists of an untrained neural network combined with an optimizer (e.g., AdamW, SGD).
When these two components are enabled with sufficient amounts of data and compute,
highly-skilled (accurate), artifacts are produced. Artifacts that possess the right abstractions for the task at hand.
The artifacts we refer to are the trained neural networks.

This points us to the idea that this paradigm, namely both the untrained NN \textbf{and} the optimizer
algorithm (as opposed to just a well-trained NN) can be what creates the novel abstractions
needed for correct predictions on the ARC private test set.
Indeed, we are the first to find success on ARC by implementing this combination of optimizer and NN in the evaluation loop.

We are able to explore what kind of training data, architecture decisions, model size, and other factors that impact test time tuning and the model's ability to create abstractions on the fly,
to solve novel ARC tasks in the forward pass.

We contribute the following:
\begin{itemize}
  \item We motivate and present architecture and pre-training recipe decisions for a performant ARC neural network in \autoref{sec:icl}.
  \item We propose the methodology for creating training data for Test-Time Fine-Tuning (TTFT) in \autoref{sec:sgd_in_eval}.
  \item We motivate and propose the Augment Inference Reverse-augmentation and Vote (AIRV) and (TTFT) as
      test-time methods for improving ARC performance (\autoref{sec:solution}),
      showing a 2.6 fold increase and a further 3 fold increase over baseline ARC-pre-training in the ARC private set accuracy respectively.
    \item
      This method helps achieve first place in the 2023 ARCathon competition,
      and achieves the highest score on the ARC private test set during the 2024 ARC kaggle competition.
      Unlike previous work, we achieve this on the completely novel ARC problems in the ARC private test set.
      We achieve the best score in the time and compute restricted kaggle test environment \cite{kaggle_arc_2020}.

\end{itemize}

\section{Problem and desired properties of a solver}
\subsection{Dataset description}

  The \emph{Abstraction and Reasoning Corpus} (ARC) dataset $\mathcal{D}$
  consists of a collection of tasks (also called riddles in this paper) ${ T_i }_{i=1}^N$, where each task $T_i$ is defined as follows:

\begin{itemize} \item A set of training examples ${ (x_j^{(i)}, y_j^{(i)}) }_{j=1}^{n_i}$,
where each $x_j^{(i)}$ and $y_j^{(i)}$ are input and output grids, for each task $T_i$ respectively. The input and output grids together are referred to as a grid-pair or example. Both terms are used interchangeably.
\item A set of test inputs ${ x_k^{(i)} }_{k=1}^{m_i}$,
with corresponding outputs ${ y_k^{(i)} }_{k=1}^{m_i}$ to be predicted.
\end{itemize}

Each grid is a 2D array $x \in C^{h \times w}$ of variable height $h$ and width $w$, where $C$ is a set of 10 colors.
The number of training examples $n_i$ and test examples $m_i$ are variable across tasks and usually range from 2-6.

The objective is to infer a task-specific function $f_i$ such that $f_i(x_j^{(i)}) = y_j^{(i)}$ for all training examples,
and then apply $f_i$ to the test inputs to obtain the test outputs.

\subsection{What does a solver need to excel on ARC?}
\label{sec:solver_need}

Each riddle is akin to a small dataset of input-to-output examples. A solver must use associative
learning as a part of the process of solving the riddle. Solvers must develop associations
between individual input and output grids and across the different input and output grid
pairs, then apply the relevant learned associations to the given test input grid.

In contrast to this, both vision-based meta-learning and natural language based reasoning datasets have a memorization problem.
High performing methods on those datasets were found to learn generic features
that allow very high levels of accuracy without significant meta-learning.
These methods were found to rely more on pretraining knowledge rather than meta-learning to gain new skills on
the new tasks provided by these datasets. \cite{mamlrapidreuse, meta_learning_without_memorization}

This is possible in those datasets because the tasks share a lot of common structure, for example
the Mini-Imagenet dataset \cite{mini_imagenet}, where the subtasks are all image-classification-based and good general
object representations can be learned and reused. This zero-shot feature reuse was shown to
take place with the MAML algorithm on Mini-ImageNet \cite{mamlrapidreuse}. These zero-shot models can outperform
meta-learning models on these tasks, holding state-of-the-art results \cite{PMF_fsl, meta_learning_without_memorization}.

For a good ARC solver, the opposite of this is desirable.
Encoding supposedly good features can lead to incorrect assumptions and missing relevant details.
Instead, it is desirable to encode a more sophisticated learning process
that can reason about the new examples and the possible transformations.
In-context-learning (ICL) is an initial candidate here \cite{gpt3, what_algorithm_is_in_context_learning}.
The ARC dataset is a challenging test of whether a system can perform this more true form of learning from a few examples.

A certain amount of flexibility and precision will be necessary within the inner workings of
a solver, to satisfy the above requirements and perform with high accuracy. The solver needs
to identify how relevant points in the input get transformed, including all the rules involved.
The solver then needs to be dynamic enough to not only develop representations, but also access
those dynamically created representations to be able to correctly apply them in the new context
(test input pair).

\subsection{The associative learning ability in the forward pass of a solver}
\label{sec:marking_problem}

Not all few-shot learning or meta-learning algorithms are well-suited for ARC. Some architectures,
such as zero-shot learners with shallow mixing, primarily rely on generating independent deep
embeddings for individual grids. These embeddings are then combined in a relatively simple
manner to produce an output grid or a classification result. One example of this is \cite{shallow_arithmetic_arc}.

One example of such an ill-suited architecture is Proto-Net \cite{protonet}. In a straightforward application
of Proto-Net to ARC, each input-output grid-pair is embedded separately into a vector, and
these vectors are averaged to create a "prototypical" representation. While this approach allows
for a simple projection-based inference, it lacks the ability to capture essential
interactions across grid-pairs

To effectively reason about transformations across multiple examples, a model needs to process
all grid-pairs in unison rather than simply averaging individual representations. Without
this capability, the model may overlook crucial shared patterns and transformations necessary
for solving the task. The limitations of such shallow interaction architectures become particularly
evident in tasks where multiple examples must be considered together to infer a general rule.

In the proto-net example model,
the only cross grid-pair interaction is the averaging of the embeddings. This
averaging is not complex enough and even may destroy information inadvertently, due
to the diversity of riddle objectives possible.

Consider a riddle that mirrors red objects horizontally and blue objects vertically. If the
test input contains both red and blue objects, a solver must see and recognize these transformations
jointly. Shallow architectures like CodeIt \cite{codeit} struggle with such tasks because they lack
mechanisms for simultaneous reasoning across all grid-pairs.

However, incorporating a structured form of associative learning in the forward pass enables
a solver to process grid-pairs holistically. This is further discussed in Section 3.1, where
we explore how structured cross-grid interactions can significantly enhance generalization
and performance.
Another example of this, applied step-by-step to an ARC riddle, is worked through in more detail in \ref{sec:sgd_in_eval}

\section{Solution}
\label{sec:solution}
\subsection{Solution Part 1: emphasizing In-Context-Learning (ICL) }
\label{sec:icl}

\subsubsection{Associative learning in LLMs' forward pass}
Recent work has uncovered growing evidence that large language models (LLMs) engage in a form
of associative learning \cite{what_algorithm_is_in_context_learning,evidence_of_meaning}. For
instance, there is now substantial support for the idea that LLMs can identify and utilize
Probabilistic Context-Free Grammars (PCFGs), effectively operating as versatile pattern-recognition
tools \cite{general_pattern_machines}. In natural language, contextual nuances play a crucial
role—each word’s meaning can shift dramatically based on its placement within a sentence—so
extensive in-context learning appears necessary for these models to generate tokens that are
both coherent and context-appropriate. Collectively, these observations indicate that LLMs
can establish and exploit relationships among input tokens, an essential ingredient for achieving
strong performance on ARC tasks.

\cite{Implicit_Representations_of_Meaning} find that models like BART and T5 can represent
and track changing entity states over the course of a narrative (e.g., whether an object is
empty or who possesses it), even without explicit supervision. Their analysis also shows that
these representations tend to be localized in specific tokens and can be directly manipulated,
causing the model to update its subsequent text generation accordingly. Crucially, most of
this dynamic-tracking ability comes from extensive open-domain pre-training rather than fine-tuning,
indicating that LLMs may possess the requisite capacity  motivated in \autoref{sec:marking_problem} for solving analogous perceptual reasoning
tasks.

\paragraph{Model choice}
We base our approach on the LongT5 encoder-decoder model \cite{longt5}, leveraging its extended context length
to accommodate larger riddles. The T5 family was selected for its sequence-to-sequence capabilities,
having been trained on a transformation task from non-causal to causal text \cite{t5}. This pre-training
instills non-causal attention mechanisms within the encoder, making it well-suited for associative
learning.

To fine-tune the model, we encode each riddle as a single text sequence, where grids
are unrolled row-wise, with pixel colors represented numerically and rows separated by spaces.
By presenting the complete riddle as a unified input, the model processes all grid-pairs simultaneously,
allowing tokens to influence each other, shown in \autoref{fig:prompt_example}. This aligns with the recommendations in Section \ref{sec:marking_problem}.

\begin{figure}[ht]
\centering
\begin{minipage}{0.95\linewidth}
\begin{verbatim}
solve: train input1 2999 4299 4442 2922 output1 19 4 4 294 2999 4429 4492 2922.
input2 27757 27525 22277 57757 52257 output2 29 5 5 275 27757 27275 27257
55727 52257.
test tinput1 4884448 8844844 4844488 4848848 8888484 8448844 8848884
toutput1 55 7 7 48 4884448 8488884 4484448 4888448 8848484 8484844 8848884.
\end{verbatim}
\end{minipage}
\caption{An example of the text prompt fed to our model. Here, each \texttt{input} and \texttt{output}
grid is unrolled into a flat sequence of pixel-color values, which are then concatenated with keywords
such as \texttt{train}, \texttt{test}, \texttt{input}, and \texttt{output}. The phrase
\texttt{solve:} indicates that the model should produce the correct transformed grid
(\texttt{toutput1}) corresponding to the given test grid (\texttt{tinput1}).}
\label{fig:prompt_example}
\end{figure}

\paragraph{Direct output}

This direct output methodology stands in contrast to some prior work that instead attempts
to produce code as an intermediate output that can then be run to produce the output grid
from the input \cite{codeit}.
While this has some benefits, such as being able to produce a solution that can be run and
tested, it also has some drawbacks.
Producing code that can solve the riddle is typically a much harder task than simply producing the output grid directly \cite{piaget1952origins_of_intelligence}.
A code based solution must be much more explicit and well defined, and it must be syntactically correct,
working correctly with all the different possible input grids.
A human solver would make a similar argument, preferring to just solve the riddle directly,
and would typically require a much shorter time to do so, compared to coding a working program that solves the riddle.

This direct output approach stands in contrast to earlier methods that generate code as an
intermediate step, which can then be executed to produce the solution grid \cite{codeit, bober2024neural}.
On the one hand, such code-based methods have the advantage of verifiability: running the program
directly tests its correctness.
On the other hand, producing code to solve a riddle is typically
more difficult than simply generating the final grid. \cite{piaget1952origins_of_intelligence}
specifying a concept or process in full detail often proves more demanding than acting out the game or task itself.
A code-based solution must be thoroughly defined,
syntactically valid, and capable of handling multiple potential input grids. By contrast, a human
solver typically finds it more straightforward and faster to solve the riddle outright
than to write, debug a general-purpose program that performs the same task.

Generating code as the final output does not fundamentally alter the broad dynamic search process
through which LLMs solve riddles—this internal flexibility and reasoning remain essential.
However, it does shift the abstraction space, training the model to handle perception and action
via programming constructs instead of direct grid outputs. A notable benefit of code generation
lies in its strong validation: one can run the generated program on the provided examples to
confirm correctness. Yet, our experiments found that the added complexity of producing a syntactically
correct, general-purpose solution introduced extra challenges and did not match the performance
of direct output generation in initial trials.

\subsubsection{Multi-task Training}

Multi-task training compels the model to manage multiple modes and contexts simultaneously,
thereby reducing its reliance on memorizing individual task details and nudging it toward genuine
contextual reasoning \cite{general_icl}. In our setup, we integrate additional tasks requiring
high levels of contextualization and reasoning—drawn from various NLP datasets—alongside the
ARC data. This approach boosts ARC performance, mirroring findings by \cite{general_icl}, who
demonstrated that vanilla transformers can exhibit robust learning-to-learn capabilities when
trained on a sufficiently large and varied set of tasks. Although \cite{general_icl} employed
simpler permuted vision tasks rather than abstraction- or reasoning-based datasets, their conclusion
that scaling task diversity helps escape the “memorization regime” remains consistent with
our observations.

\subsubsection{Code pre-training and contextualization}
We observe that training on coding tasks offers a more pronounced performance boost on ARC
than merely adding multi-task training derived from language or NLP domains.
It is generally easier to continue a sentence halfway through a paragraph than a code file halfway.
Code datasets inherently demand meticulous attention to detail and context, requiring the model to keep track
of variables and resolve dependencies.
This greater focus on accuracy and hierarchical reasoning means that memorization alone is insufficient—an important
distinction from many NLP tasks where world knowledge and memorized associations play a
larger role. As noted by \cite{code_which_stage}, coding data “is more logical and less ambiguous”
ultimately fostering a better focus on context.

A more complete discussion of recent literature supporting the use of code data in LLM training can be found in \autoref{sec:literature_review}.

\subsubsection{Automatic Riddle Generators}
\label{sec:riddle_fairness}

Programmatic riddle generation is a valuable strategy to expand ARC training data and enhance
model learning. To facilitate this, we employ Domain Specific Language (DSL) techniques, drawing
inspiration from the work of Andreas Koepf \cite{Koepf2022} and Johan Sokrates Wind’s (Icecuber)
DSL \cite{ice_cuber}, to construct synthetic riddles by sampling function names and their parameters. A
key aspect of our training approach involves training the model to infer these underlying DSL
function names and parameters from the input riddle grids, in addition to predicting the final
output grids. This dual-prediction strategy, where models learn to predict both the output
grid and the DSL function names, contributes to more robust performance compared to training
on either target alone. While DSL-generated data represents a portion of our overall training
corpus, it serves to illustrate important data generation concepts we utilize.

To further diversify our ARC-specific training data, we also employ various more traditional
riddle generators. These generators produce complete input-output pairs and test grids, enriching
our training dataset. We observe that if riddle examples leave aspects of the transformation
underspecified, the model may inadvertently learn to encode these ambiguities directly into
its weights, rather than inferring them contextually. This can lead to undesirable biases and
reliance on memorization. To mitigate this, we deliberately err on the side of overspecification
in our generated riddles, ensuring sufficient information for the model to unambiguously determine
the intended solution.

We give a more detailed description of the synthetic riddle generators in Appendix A.

\subsection{Solution Part 2: Optimizing in the Evaluation Loop (Test-Time Fine Tuning)}
\label{sec:sgd_in_eval}
During evaluation, we leverage each test riddle’s demonstration examples to create synthetic
training data.
Specifically, we select a grid pair from the riddle’s provided examples and
repurpose it as a new “test example” forming a new, smaller riddle. We have the answer to this
new riddle, as it was taken from the demonstration examples, and so we can use that to train the model at test time.
To get a lot more data and
ensure this riddle differs from the original, we apply several augmentations:
\begin{itemize}
    \item \textbf{Color permutation:} Randomly shuffle the color labels throughout the riddle.
    \item \textbf{Spatial transformations:} Rotate, flip, or transpose the input and output
grids, sampling from the dihedral group $D_{4}$.
    \item \textbf{Shuffling:} Randomly reorder the input demonstration examples.
\end{itemize}
We then perform a brief round of fine-tuning on these augmented riddles before generating predictions
for the test grids. This procedure can be seen as a form of test-time training \cite{sun2020testtime}
and is referred to here as \textit{Test-Time Fine Tuning} (TTFT).

\subsubsection{Motivation for TTFT Through Iterative Reframing}
\label{sec:motivation_ttft}

\begin{wrapfigure}{H}{0.7\textwidth}

  \vspace{-15pt} %
  \centering
  \includegraphics[width=0.5\textwidth]{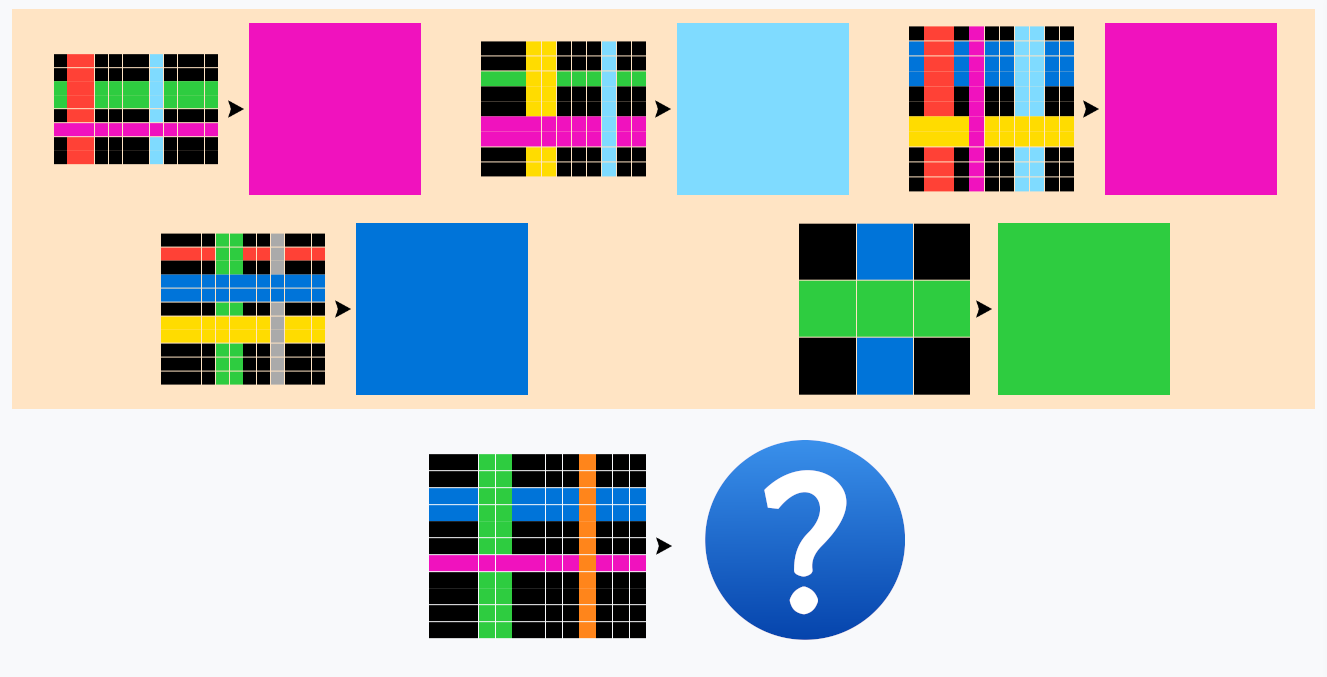}
  \caption{An example of a simpler ARC riddle.}
  \label{fig:arc_example_2}
  \vspace{-15pt} %
\end{wrapfigure}

The central idea behind Test-Time Fine Tuning (TTFT) is that the solver may initially make
mistakes on the private test set, and we can exploit that feedback to refine its approach.
Much like a human solver, the model can re-evaluate and iterate on potential solutions. For
instance, consider the scenario illustrated in Figure~\ref{fig:arc_example_2}, where the correct
transformation is to select the color of the line that does not intersect others. A solver
or human might instead begin by hypothesizing that the intended solution is to choose the color
of the \emph{thinnest} line, perhaps because the first few examples happen to support that
interpretation.

This kind of oversight often arises from limited attention or inadequate depth of processing: the model (or a human) fails to fully observe all relevant information and thereby locks into a flawed “framing.” Once the solver is committed to an incorrect framing (for example, consistently searching for the thinnest line), it will have to continue to discard potentially crucial data and remain blind to the correct pattern.
The input grids have already been processed with the incorrect framing/bias.
In such a case, only a complete reprocessing of the grids—where the correct framing is established from the outset—can guide the solver to correctly infer the solution.

Proper framing is consistently shown to be critical for perceptual models, greatly influencing
performance. One illustration of this appears in \cite{lost_in_the_middle_attention}, which
observes that language models often ignore information in the middle of a prompt, yet perform
significantly better when the framing (in the form of the question or key instruction) appears
at both the beginning and end of the prompt.
Another relevant example is instruction tuning in LLMs \cite{ouyang2022traininglanguagemodelsfollow,flan},
wherein models are trained to adopt a “helpful” framing instead of a purely next-word prediction
mode.

Hence, mechanisms for “reframing” are crucial in solving perceptually rich tasks, especially
in the ARC setting, where each riddle’s solution demands a tailored framing. TTFT provides
a means to adapt these frames based on feedback derived from newly generated training data—mirroring
the human process of iteratively revisiting and adjusting hypotheses until the training examples
are correctly solved, before finally addressing the test grids.

\paragraph{Why take full parameter update steps (Full fine tuning)}
Although several lighter-weight alternatives exist for adapting models to downstream tasks—including
chain-of-thought prompting \cite{cot_prompting}, few-shot prompting \cite{gpt_3}, and low-rank
adaptation methods \cite{lora_first}—we chose full parameter updates primarily due to simplicity
and reliability. Training the model using full parameter updates is naturally powerful enough
to generate the needed abstractions. Given ARC's demand for generating diverse and genuinely
new abstractions at test-time, we opted for this straightforward, guaranteed method of update,
even though other adaptation techniques may also offer promising results and sufficient updating
power.

\subsubsection{Attention and masking}
We specifically choose encoder-decoder architectures because they incorporate non-causal (unmasked)
attention within the encoder, allowing each token to simultaneously attend to the entire input
sequence. This capability is critical for enabling the model to fully interpret and contextualize
ARC riddles from the outset.

By contrast, if the riddle were presented using causal (masked) attention, tokens appearing
earlier in the sequence would not have access to the complete context, preventing them from
forming accurate early-stage representations or hypotheses simply due to lack of available
information. Tokens representing input grids would be unable to attend forward to their corresponding
output grids, significantly limiting the model’s reasoning about intended transformations.
To validate the practical importance of this non-causal attention mechanism, we experimentally
compared our encoder-decoder approach against similarly sized causal decoder-only models and
found that the encoder-decoder structure yielded substantially better performance.
We were not able to run experiments to disambiguate whether this is due to the non-causal attention masking or the encoder-decoder architecture itself,
but its likely that the non-causal attention is the main factor here.

\subsubsection{Specialization to the riddle}
Test-time fine-tuning can also enhance the precision required to produce completely accurate
outputs. Even when the model correctly identifies the transformation function, minor execution
errors—such as inaccuracies of a pixel or two—may occur. These errors are likely due to limited
model depth or capacity, restricting the model's ability to execute transformations perfectly
on the first attempt. TTFT can mitigate these issues by adapting the model specifically to
the current riddle, refining its "execution" capabilities, and enabling it to achieve precise,
pixel-perfect outputs.

\subsubsection{Beam Search for Solution Space decoding}

Because our model generates output grids autoregressively, a purely greedy decoding strategy
is brittle: even a single incorrect token leads to an unrecoverable trajectory. Beam search
\cite{beam_search} addresses this by maintaining multiple candidate solutions simultaneously,
pruning all but the most promising branches at each decoding step according to cumulative probabilities.

This strategy allows the solver to effectively handle cases where the correct next token may
initially have a lower probability but becomes clearer in subsequent steps. Conversely, incorrect
trajectories naturally lose confidence as they proceed; a well-calibrated model will assign
increasingly uniform probabilities across candidate tokens when uncertain, causing these erroneous
paths to be rapidly discarded.
Although models can occasionally become \emph{confidently} wrong, beam search generally remains
beneficial \cite{beam_search}, and the ARC dataset in particular reaps substantial advantages
from this capability: each riddle has precisely one correct solution, amplifying the divergence
between correct and incorrect paths under beam search.

\subsection{Augment, Inference, Reverse augmentation and Vote (AIRV)}
\label{sec:airv}

\begin{figure}[t]
    \centering
    \includegraphics[width=\linewidth]{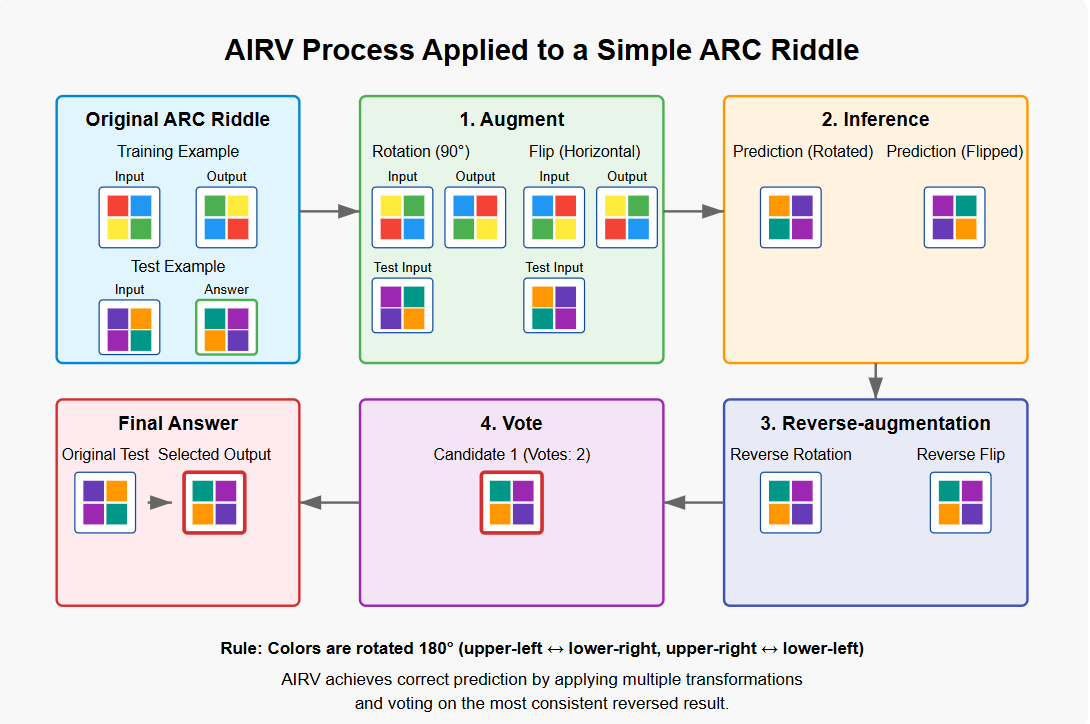}
    \caption{
        \textbf{AIRV process applied to a simple ARC riddle.}
        Starting from the original riddle (blue panel), the pipeline
        \emph{(1) Augments} the grids via rotations and flips,
        \emph{(2) runs inference} on each transformed instance,
        \emph{(3) reverses} every prediction back to the original frame of reference,
        and finally \emph{(4) votes} on the most consistent output.
    }
    \label{fig:airv_process}
\end{figure}

We propose a test-time augmentation strategy called \textit{Augment, Inference, Reverse-Augmentation,
and Vote (AIRV)}. The procedure begins by applying a spatial transformation to the input riddle
(e.g., rotation or flipping). We then get predictions on the transformed riddle to obtain a
predicted output grid, which is subsequently \textit{reversed} back to the original orientation.
Finally, we gather multiple such predictions from different spatial augmentations and use a
voting scheme to select the most frequent (or most confident) output grid.

Unlike beam search \cite{beam_search} or temperature sampling \cite{temperature_effect}, AIRV
can generate duplicate predictions (after reversing). This enables a voting mechanism that amplifies
strong, consistent solutions and filters out noisy variants.
This is particularly useful in the ARC setting, where each riddle has only one correct answer.
Assuming a somewhat competent model,
a voting mechanism then provides a very effective way to
make salient the more dominant and consistent grid predictions
(the more dominant solution ideas) from other more noisy predictions.
In our opinion, this works because given a reasonably performant model, there are many ways that
solutions can be incorrect, while there is only exactly one correct solution.
This can be seen as analogous to the clustering step in the AlphaCode methodology \cite{alpha_code}
where the generated programs are clustered and the most common program cluster is selected as a proxy
for most likely correct program.

\section{Results}

\subsection{ARC dataset splits}

The ARC dataset consists of 400 training riddles, 400 public evaluation riddles, and 100 private evaluation riddles that are not accessible to the public \cite{arc_agi_2024}.
The training set riddles are the easiest, then public evaluation riddles are harder, and finally the private test set have been
shown to be harder than the public evaluation set \cite{Knoop_2024}.

\subsubsection{Testing setup}

We report our results on the private test set, with the following test-time compute limitations imposed by the
competition compute environments available \cite{kaggle_arc_2020, arc_agi_2024}:
Namely 2 hours of runtime on a single P100 GPU (16 GB VRAM).

Accuracy between predicted output grids and the ground truth output grids
is measured by only counting exact matches over the whole predicted grid,
allowing for two grid attempts per task (top-2).

\begin{figure}[h]
  \centering
  \includegraphics[width=0.4\textwidth]{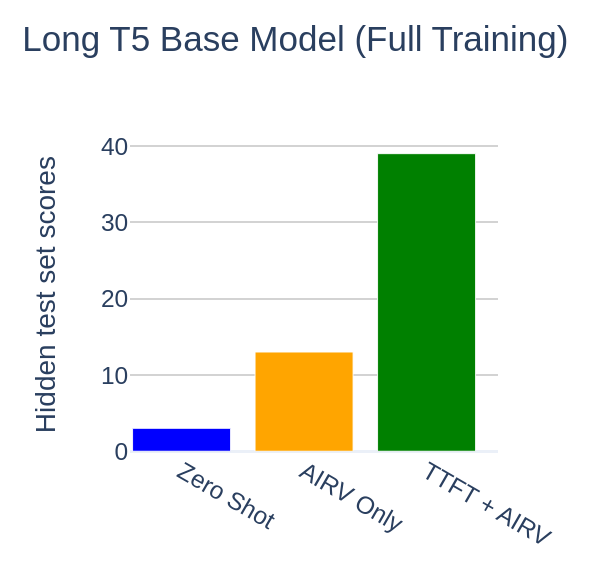}
  \caption{Results for the fully trained base model in the 3 different test-time configurations.}
  \label{fig:base_model_score}
    \centering
    \subfloat[\centering Zero-Shot]{{\includegraphics[width=0.29\textwidth]{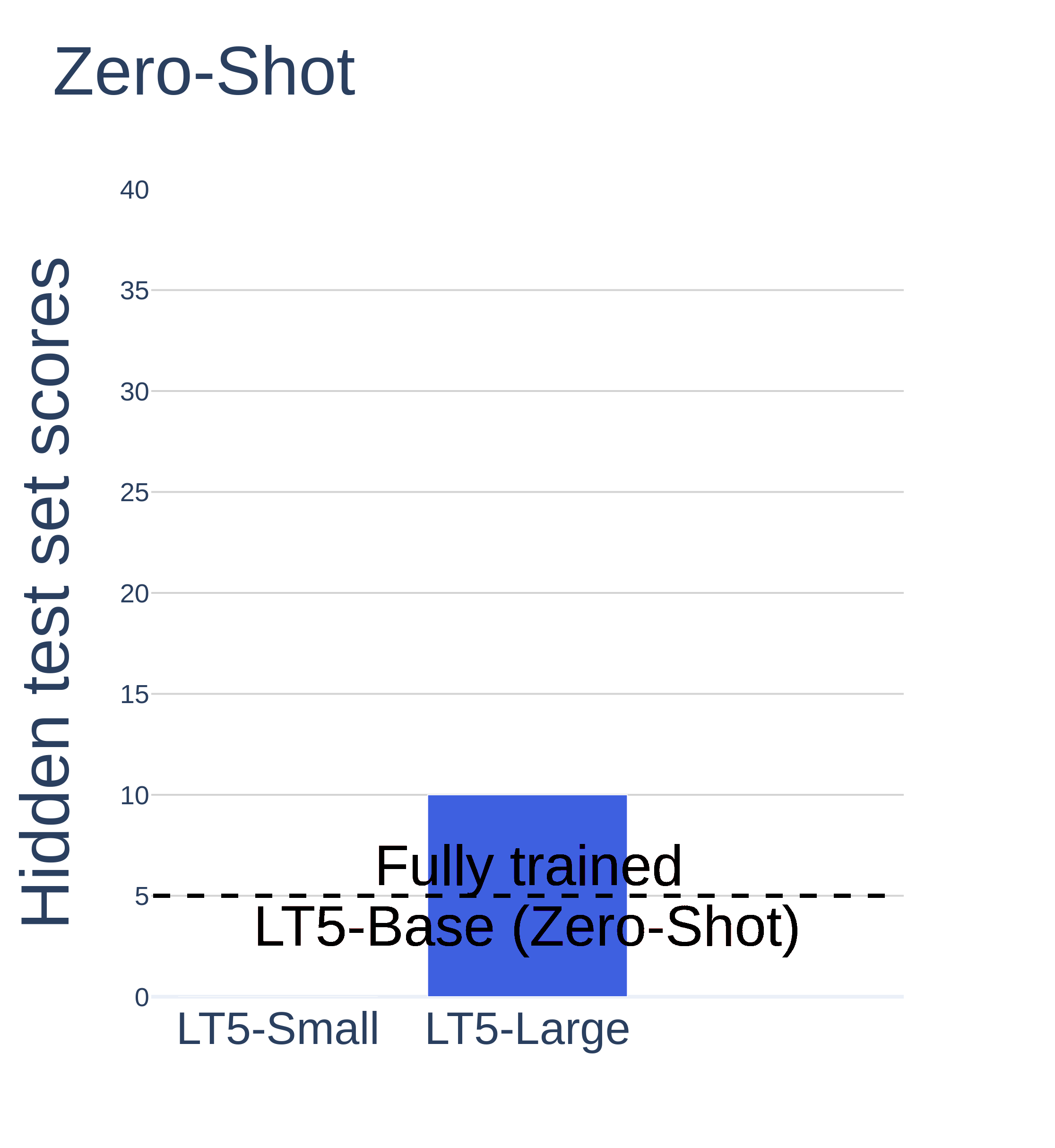} }}%
    \qquad
    \subfloat[\centering AIRV only]{{\includegraphics[width=0.29\textwidth]{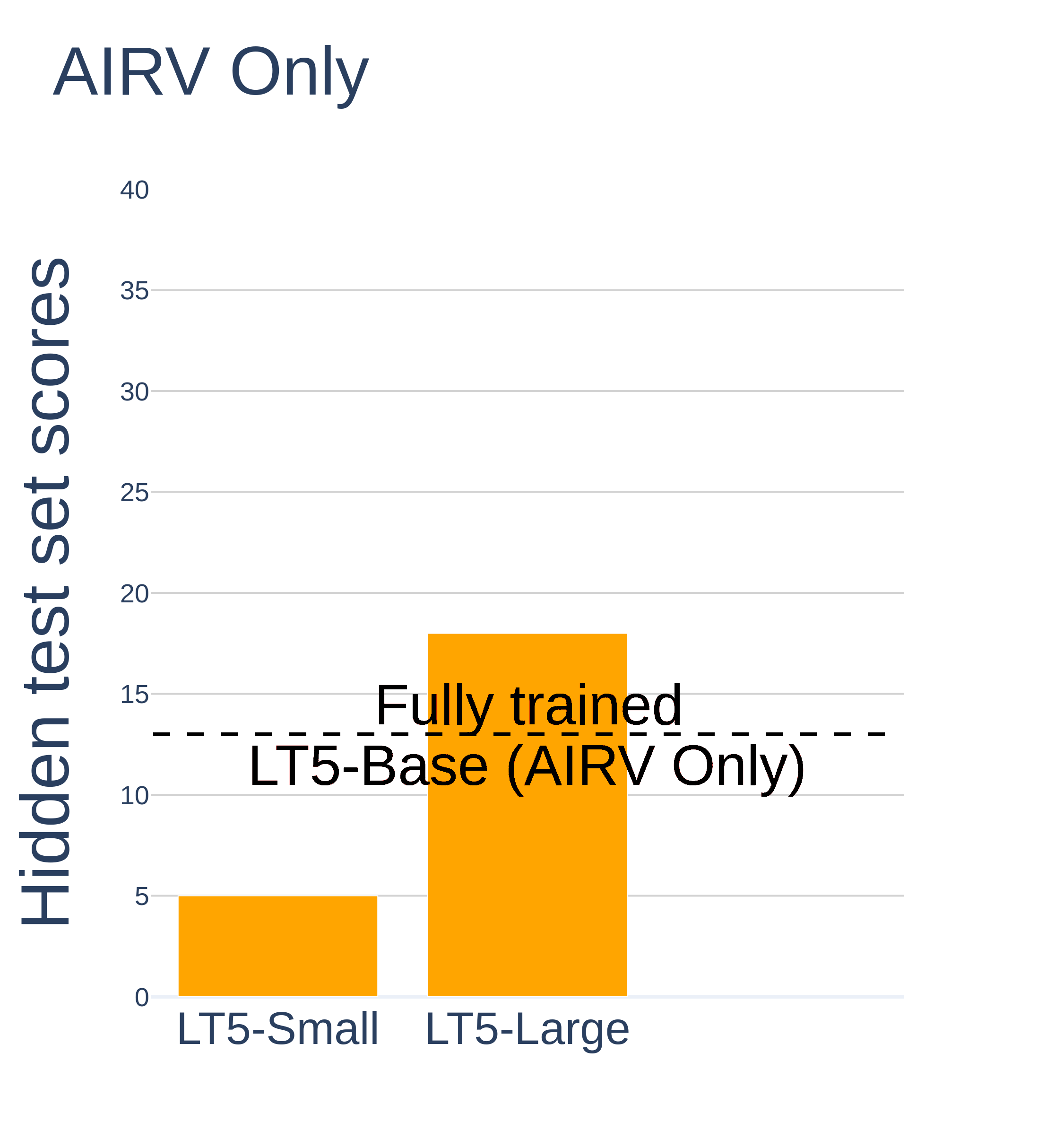} }}%
    \qquad
    \subfloat[\centering TTFT + AIRV]{{\includegraphics[width=0.29\textwidth]{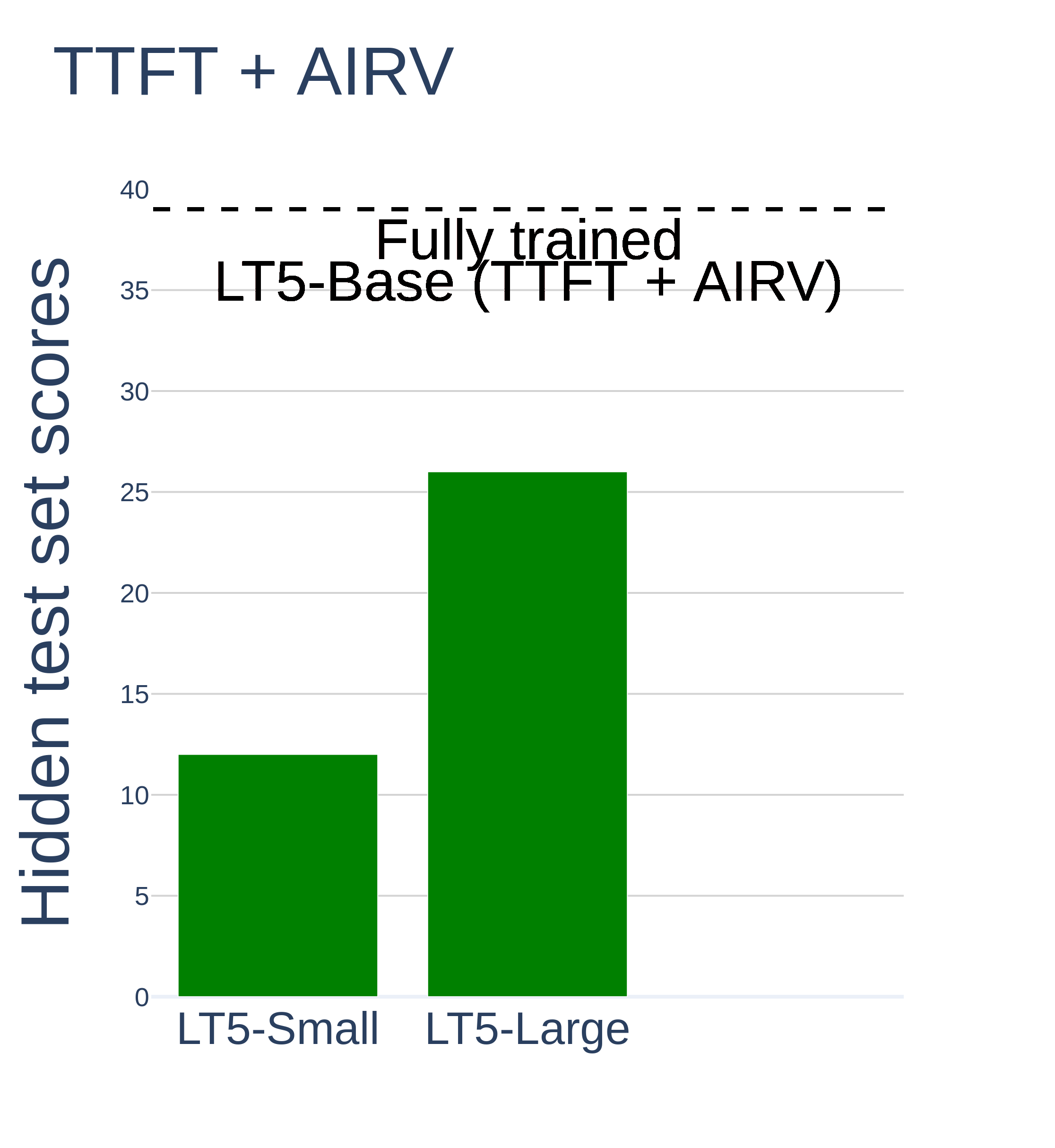} }}%
    \caption{Performance on the ARC's private test set. Each subfigure analyzes the impact of
      the different configurations of the test time techniques introduced.
      The small and large models trained on a subset of the training data are compared to the
      base model trained on the full training data.
      The effect of increasing model size is contrasted with the effect of increasing the number
      of training examples across the different test time techniques.
      We see that, model size has a significant impact despite a significantly reduced training set,
      except for when TTFT implemented at test time.
    }
    \label{fig:example}%
\end{figure}

\subsection{Analysis and Discussion}
\label{sec:analysis_performance}
To evaluate the effectiveness of our methodology, we compare the following configurations:

\begin{itemize}
    \item \textbf{Zero Shot (No TTFT/AIRV):} Direct prediction using the ARC-trained LongT5 model.
    \item \textbf{AIRV Only:} Applying the AIRV technique (with beam search decoding) but without TTFT.
    \item \textbf{TTFT + AIRV:} Combined use of TTFT and AIRV (with beam search decoding).
\end{itemize}

We also train small and Large LongT5 variants on our training data. Due to pre-training compute constraints,
only train those models on around 10\% of the total training data.

\begin{table}[h]
  \centering
  \begin{tabular}{|c|c|c|c|}
    \hline
    \textbf{Model} & \textbf{Zero Shot (No TTFT/AIRV)} & \textbf{AIRV Only} & \textbf{TTFT + AIRV} \\
    \hline
    Fully trained Base LongT5 & 5 & 13 & 39\footnote{3} \\ %
    \hline
    Partially trained small LongT5 & 0 & 5 & 12 \\
    Partially trained large LongT5 & 10 & 18 & 26 \\
    \hline

  \end{tabular}
  \end{table}

\footnotetext{A version of this approach with more optimizations and extended ARC training, achieved the highest score on the ARC-AGI private test set in 2024 (58\%) \cite{arc_agi_website}.}

\paragraph{AIRV:} Augmentation and voting also enable higher performance. Absolute AIRV gains show positive scaling
with model size. AIRV alone can gain up to 260\% (in the base model, fully trained scenario).

\paragraph{TTFT:} Test-time fine-tuning significantly increases the model’s score,
results are consistent across model sizes and training regimes. Performing TTFT before running inference with AIRV leads to an additional 300\% gain in performance.

An early version of this approach achieved first place in the 2023 ARCathon \cite{arcathon2023}.
A version of this approach with more optimizations and extended ARC training, achieved the highest score on the ARC-AGI private test set in 2024 (58\%) \cite{arc_agi_website}.

\paragraph{Experiments on model size}

The Large model gets a lower score on TTFT but a higher score on Zero Shot and AIRV only, even with the lighter training.
We see this trend holding across small, base and large models, where zero shot and AIRV only performance
trends with size of the model.
This scaling behavior is typical in deep learning \cite{Kaplan2020ScalingLF,chain_of_thought_prompting}, but in this testing regime (ARC), can be explained by the fact that
the larger the model the bigger and more expressive forward pass will be (more, and wider layers means more associations that can be made in the forward pass).
This possibly accounts for
the increased performance in the
AIRV and zero shot setting,
aligning well with the motivation in \autoref{sec:icl}.

Benefits of increased pre-training on the ARC dataset did not beat the benefits of simply increasing the model size.
This perhaps indicates that the forward pass flexibility of the models is impacted much more by the model size
compared to pre-training. This aligns with general theory regarding model scaling laws and how they impact typical
reasoning benchmarks \cite{chain_of_thought_prompting, gpt3}.

\paragraph{The effect of pre-training on TTFT}
Interestingly, while increased pre-training does not seem to beat the effect of simply increasing the model size,
when it comes to zero shot and AIRV only performance of these models, it does significantly improve score in the
TTFT + AIRV setting compared to the base model. This effect is not likely only due to that larger models take more time to train (and TTFT)
as the base model still sees a much larger boost in performance (300\%) from TTFT than \emph{both} large and small models (140\% and 240\% respectively).
We discuss why this may be in \autoref{sec:ctx_is_important}.

\paragraph{Contextualization at Test Time vs.\ Pre-Training on ARC Riddles}
\label{sec:ctx_is_important}

We have motivated why high quality contextualization is a crucial element for tackling the
ARC dataset. Yet, our experimentation shows that substantial pre-training on ARC riddles remains
indispensable for achieving state-of-the-art performance with TTFT, culminating in our 2024 highest score
on ARC-AGI.
While the space of possible ARC transformations is vast, pre-training does not merely “leak”
memorized solutions. Instead, it imparts both the foundational “core knowledge priors” described
by \cite{Chollet2019OnTM} and a range of \emph{more subtle but highly important} priors. By
this, we refer to heuristics such as a preference for simpler, human preferred transformations
(the “simple/simpler transformation” bias), a drive to validate transformation hypotheses across
examples (the “looking for confirmation” bias), and even the basic notion that each example
is formed by a paired input and output grid. Without these biases deeply embedded in the model
weights, the solver would be far less efficient in forming or testing hypotheses during inference,
and a lot of the forward pass would be spent at just merely realizing these basic things about the problem setup.

This is further supported by the following recent findings.
\cite{hermann2023foundations,shah2020pitfalls} have demonstrated that predictive features emerge
at different layers during pre-training, with simpler but equally predictive features appearing
earlier in the network.
Recently, \cite{sedementation_paper} also show that longer pre-training allows for complex
but predictive representations to "sediment" (move  into the earlier layers).
We hypothesize that the extensive ARC pre-training allows for more "room" for test time features
to emerge (during test-time fine-tuning), because the base arc priors have sufficiently sedimented
into the very early layers.
This sedimentation process, may be the crucial key for enabling the model to handle more complex,
task-specific reasoning when test-time fine-tuned on unseen riddles.

\paragraph{Contrasting pre-training with program synthesis}
These priors are particularly relevant when considering a solver that is based on program synthesis.
Program-synthesis-based approaches explicitly encode pair association and transformation-confirmation
heuristics by searching for a program that correctly transforms input grids into their corresponding
output.
While these methods avoids the need for extensive domain-specific pre-training, it still
requires significant human intervention to guide the program synthesis algorithm.
Specifically, humans must direct the algorithm to search for transformations that align input grids with
output grids and ensure that the transformations are correct using the other grid-pairs.
Moreover, program synthesis solvers, with their manually encoded heuristics, are generally
less effective when faced with perceptual problems that involve an almost limitless range of
possible transformations and framings.

These are important considerations to keep in mind when considering the trade-offs between
extensive pre-training compute and why its necessary, and the explicit program synthesis approach.

\section{Related work}
\label{sec:literature_review}

Classically, meta-learning can be regarded as a strategy to automate model design or parameter
selection across numerous tasks, often formulated as a two-level optimization problem. In such
setups, an “outer” model accumulates meta-knowledge, while an “inner” model adapts rapidly
to each new task. More recent advances in \emph{in-context learning} (ICL) have sidestepped
explicit inner–outer distinctions, instead relying on the model’s forward pass to perform meta-learning.
This behavior is commonly observed in transformer architectures trained on data with specific
distributional properties \cite{data_distributional_properties_ICL}, enabling impressive performance
on various tasks. ICL appears to support a more data-efficient form of meta-learning, one capable
of storing and leveraging priors in a flexible way—especially appealing for applications like
ARC.

A notable illustration of forward-pass ICL exhibiting strong generalization is the MLC architecture \cite{Lake2023Humanlike}.
MLC mines for examples similar to the new task on their dataset, and, similar to our methodology, combines all relevant examples into the model’s forward pass at once, allowing it to function as a meta-learner \emph{within} the forward pass. This design substantially improves performance and generalization, highlighting the potential of ICL-based approaches for tasks that require complex reasoning.

\subsection{Concurrent Work Building on TTFT}
An interesting replication that is based on our proposed Test-Time Fine Tuning (TTFT) approach
can be found in the work of \cite{the_surp}, who explore the idea of adapting models on-the-fly
for ARC tasks. Their method explicitly incorporates a similar short fine-tuning phase during
inference on ARC, closely mirroring our TTFT paradigm. This aligns with our findings that dynamic
adjustments at evaluation can significantly enhance performance, especially when encountering
tasks requiring newly discovered transformations or abstractions.
Also based on \cite{community_post, mlst_ep} their methodology and results strongly corroborate
and align with ours.

Building further upon our proposed TTFT approach, \cite{transduction_induction} recently explored
the interplay between inductive and transductive reasoning specifically within the ARC domain.
Their study trains neural networks on synthetic datasets generated from Python-based implementations
of ARC transformations, based on \cite{community_post}, they also use TTFT for their transductive
domain. Their experiments highlight that inductive program synthesis excels in precise symbolic
tasks, whereas transduction demonstrates strength in more perceptually oriented problems. By
effectively ensembling these complementary models, their approach achieves strong results, strongly
validating the effectiveness and flexibility of TTFT-based adaptation to achieve high performance.

\subsection{Code data in LLM training}

Emphasizing coding and code training in LLMs is not new.
Coding datasets form a significant part of LLM pre-training corpora, even in non-coding
based models \cite{palm,touvron2023llama}.
It is correlated with improved performance on reasoning tasks, \cite{HELM_full_eval} find that code based
models consistently outperform text based models in reasoning,
even on synthetic reasoning tasks formulated as natural text.
Further, \cite{code_which_stage} show that pre-training LLMs with a mix of text and code increases the general reasoning capability of the model.
They also show that code at the instruction tuning stage enhances task specific reasoning.
\cite{code_commonsense_few_shot} also show that code models, even when outputting text, outperform text models on few shot structured reasoning evaluations.
More recently, \cite{code_ablate} carefully ablate the effects of code-data in pre-training and find positive effects on compositional tasks like semantic parsing and mathematics.

\subsection{ARC dataset's related work}

The Abstraction and Reasoning Corpus (ARC) dataset \cite{Chollet2019OnTM} presents a significant
challenge for artificial intelligence systems due to its emphasis on reasoning from minimal examples.
Numerous approaches have been proposed to tackle ARC tasks,
ranging from leveraging LLMs to developing specialized neural architectures
and neuro-symbolic methods to brute force search based methods.

\subsubsection{Evaluating LLMs on ARC Tasks}

Several studies have explored the capabilities of LLMs on ARC tasks without
additional training.
Mirchandani et al.\cite{mirchandani2023large} investigated whether LLMs can act as general
pattern machines by providing the entire ARC task as context to GPT-4 and GPT-3.5. They achieved
an accuracy of 10.6\% on the combined ARC dataset and 6.75\% on the public test set, indicating
limited performance. Similarly, Mitchell et al.\cite{mitchell2023comparing} compared GPT-4
and GPT-4V to human performance on a simplified version of ARC called ConceptARC \cite{moskvichev2023the},
finding that GPT-4V did not significantly improve performance over GPT-4 and that both models
underperformed compared to humans.

Other works have attempted to improve LLM performance by altering input representations.
\cite{10208934} translated visual ARC tasks into textual descriptions
to leverage LLMs' reasoning capabilities, achieving 20\% accuracy with GPT-3 on the ARC training
set. \cite{xu2024llms} emphasize the importance of object-based representations,
introducing an object-centric encoding to feed into LLMs. They tested GPT-4 on the easiest
50 ARC tasks and solved 23 out of 50.

These studies highlight that frozen LLMs possess some pattern recognition abilities and may possess some associative learning abilities
out of the box in their pretrained forward pass,
but they struggle with the abstraction and reasoning required for ARC tasks.

\subsubsection{Neuro-Symbolic and Program Synthesis Approaches}

Another line of research focuses on combining neural networks with symbolic reasoning or program
synthesis to solve ARC tasks. Wang et al.\cite{wang2024hypothesis} proposed a method where
GPT-4 generates high-level textual hypotheses about the task, then translates these into code
to solve the task. Testing on a subset of 40 ARC training tasks, they achieved a success rate
of 27.5\%, which increased to 37.5\% with human selection of correct hypotheses. However, this again
relies heavily on the frozen LLM's forward pass to be powerful enough to do the reasoning required for ARC tasks.

\cite{butt2024codeit} introduced CodeIt, a method that generates code based on
grid-pairs and uses a self-improving loop during evaluation. CodeIt solves 14.8\% of the ARC
evaluation tasks and runs into the limitations we describe in \ref{sec:marking_problem}.

\cite{lei2024generalized} developed Generalized Planning for ARC
(GPAR),
modeling ARC tasks as generalized planning problems in the Planning Domain Definition Language (PDDL)
coupled with external functions representing object-centric abstractions of the grids.
\cite{lei2024generalized} achieved 50\% accuracy on a subset of object-centric tasks.
\cite{bober2024neural} used a dreamcoder inspired approach \cite{dreamcoder} with a domain-specific language for ARC
tasks, achieving 18 out of 400 tasks on the ARC evaluation set.

These highlighted approaches attempt to incorporate a more symbolic reasoning approach to
solve ARC tasks yet can only achieve a limited success rate or work only within a specialized domain.
This may be due to the limitations of perceptual reasoning with these symbolic approaches, or may also
be in some cases due to the added complexity of fully representing ARC transformations in code or domain-specific languages.

\cite{shallow_arithmetic_arc} utilized neural embeddings and vector arithmetic to solve ARC visual analogies but achieved
only 2\% accuracy on the public evaluation set. We discussed this weaknesses of this approach further in \ref{sec:marking_problem}.

\subsubsection{Brute Force Search}

Icecuber \cite{ice_cuber} attempts to solve ARC by performing a brute force search over unary
functions on pieces of input grids, forming a directed acyclic graph (DAG) of many possible
transformations, until a DAG that creates the training output grid is found.
This method won the 2020 ARC challenge competition on Kaggle \cite{kaggle_arc_2020}.

\subsubsection{Other Datasets}

Other works have focused on simplified versions of ARC. \cite{assouel2022objectcentric} state that ARC is
impenetrable for the time being and introduce the Sort-of-ARC dataset, which is only limited to 20x20 grids and
only contains 3 objects with a limited set of transformations only. They emphasize object centric reasoning by
using a controller to generate a solution vector, use slot attention transformer to extract
object vectors, then update the solution and object vectors in an object centric way before
generating the final solution using a spatial decoder on the result. We believe that a purely object centric
approach does not generalize to tasks where objects are ambiguous, and the correct "object" is very task specific.
\cite{assouel2022objectcentric} achieved 59\% accuracy on out-of-distribution tasks in the Sort-of-ARC dataset.

\cite{xu2024llms} introduce the 1D ARC dataset, which is a simplified version of ARC with only one dimension.
They emphasize the importance of object centric input representations and prompt GPT-4 with their
object centric representations of the tasks. They state that they strategically select the easiest 50
tasks out of the training set and solve 23 out of the 50 tasks.

\subsubsection{Summary and limitations of related work}

To summarize, ARC has indeed proven to be a challenging problem for the current paradigm of AI,
with state-of-the-art results remaining low on the private test set compared to other datasets in the field,
despite 3 recent competitions on the ARC \cite{kaggle_arc_2020,arcathon2023}.

We identify the following limitations of previous approaches:
\begin{itemize}
  \item \textbf{Comparisons without computational constraints:} Some studies compare their
    score to others without reporting the computational cost of their methods, making it difficult to
    make an apples to apples comparison. We rather rely on private test set performance and the computational constraints set by kaggle to avoid this issue.
    This is a major weakness of related work, we find that our method is much more computate efficient while still achieving state-of-the-art.
    \item \textbf{Limited Generalization:} Many methods perform well on subsets of ARC tasks
or simplified versions but fail to generalize across the full spectrum of ARC tasks.
    \item \textbf{Lack of Contextual Reasoning:} Approaches that process grid-pairs in isolation,
      such as CodeIt \cite{butt2024codeit} and \cite{shallow_arithmetic_arc} and GPAR \cite{lei2024generalized}
struggle with tasks that require understanding relationships across multiple examples.
     \item \textbf{Lack of focus on perceptual reasoning:} Methods that do not focus on perceptual reasoning seem to face
 difficulties due to the complexity of searching through an almost infinite space of possible transformations.
    \item \textbf{Evaluation on Public Data:} Some studies evaluate their models on the public
ARC dataset, which may have been exposed in pre-training data, potentially inflating performance
estimates.
\end{itemize}

\section{Conclusion}

In contrast to previous work, our approach leverages LLMs with a focus on tackling arc with broader
perceptual reasoning as the core component. We achieve this by both training a performant
perceptual reasoner using existing state-of-the-art contextualizing models (LLMs).
And secondly, by fine-tuning this perceptual reasoner on ARC tasks for each task at test time.
Unlike previous methods we prove out our
approach by evaluating on the full private test set,
ensuring that our methodology generalizes well to unseen tasks.
We achieve state-of-the-art performance on the ARC-AGI private test set while being much more computationally efficient than existing methods.

\section*{Acknowledgments}
This research was conducted with the help of the Google TPU Research Cloud (TRC) program.
We would like to thank the TRC program for providing the TPU resources and for their continued support.

\bibliographystyle{unsrt}
\bibliography{references}

\begin{thebibliography}{10}

\bibitem{Chollet2019OnTM}
François Chollet.
\newblock On the measure of intelligence.
\newblock {\em ArXiv}, abs/1911.01547, 2019.

\bibitem{AreWeDoneWithImageNet}
Lucas Beyer, Olivier~J. H'enaff, Alexander Kolesnikov, Xiaohua Zhai, and A{\"a}ron van~den Oord.
\newblock Are we done with imagenet?
\newblock {\em ArXiv}, abs/2006.07159, 2020.

\bibitem{concept_arc_benchmark}
Arsenii~Kirillovich Moskvichev, Victor~Vikram Odouard, and Melanie Mitchell.
\newblock The concept{ARC} benchmark: Evaluating understanding and generalization in the {ARC} domain.
\newblock {\em Transactions on Machine Learning Research}, 2023.

\bibitem{general_pattern_machines}
Suvir Mirchandani, Fei Xia, Pete Florence, brian ichter, Danny Driess, Montserrat~Gonzalez Arenas, Kanishka Rao, Dorsa Sadigh, and Andy Zeng.
\newblock Large language models as general pattern machines.
\newblock In {\em 7th Annual Conference on Robot Learning}, 2023.

\bibitem{Devlin2019BERTPO}
Jacob Devlin, Ming-Wei Chang, Kenton Lee, and Kristina Toutanova.
\newblock Bert: Pre-training of deep bidirectional transformers for language understanding.
\newblock In {\em North American Chapter of the Association for Computational Linguistics}, 2019.

\bibitem{VIT}
Alexey Dosovitskiy, Lucas Beyer, Alexander Kolesnikov, Dirk Weissenborn, Xiaohua Zhai, Thomas Unterthiner, Mostafa Dehghani, Matthias Minderer, Georg Heigold, Sylvain Gelly, Jakob Uszkoreit, and Neil Houlsby.
\newblock An image is worth 16x16 words: Transformers for image recognition at scale.
\newblock In {\em International Conference on Learning Representations}, 2021.

\bibitem{kaggle_arc_2020}
François Chollet, Katherine Tong, Walter Reade, and Julia Elliott.
\newblock Abstraction and reasoning challenge.
\newblock \url{https://kaggle.com/competitions/abstraction-and-reasoning-challenge}, 2020.
\newblock Kaggle.

\bibitem{mamlrapidreuse}
Aniruddh Raghu, Maithra Raghu, Samy Bengio, and Oriol Vinyals.
\newblock Rapid learning or feature reuse? towards understanding the effectiveness of maml.
\newblock In {\em International Conference on Learning Representations}, 2020.

\bibitem{meta_learning_without_memorization}
Mingzhang Yin, George Tucker, Mingyuan Zhou, Sergey Levine, and Chelsea Finn.
\newblock Meta-learning without memorization.
\newblock In {\em International Conference on Learning Representations}, 2020.

\bibitem{mini_imagenet}
Oriol Vinyals, Charles Blundell, Timothy Lillicrap, koray kavukcuoglu, and Daan Wierstra.
\newblock Matching networks for one shot learning.
\newblock In D.~Lee, M.~Sugiyama, U.~Luxburg, I.~Guyon, and R.~Garnett, editors, {\em Advances in Neural Information Processing Systems}, volume~29. Curran Associates, Inc., 2016.

\bibitem{PMF_fsl}
Shell~Xu Hu, Da~Li, Jan Stühmer, Minyoung Kim, and Timothy~M. Hospedales.
\newblock Pushing the limits of simple pipelines for few-shot learning: External data and fine-tuning make a difference.
\newblock In {\em 2022 IEEE/CVF Conference on Computer Vision and Pattern Recognition (CVPR)}, pages 9058--9067, 2022.

\bibitem{gpt3}
Tom Brown, Benjamin Mann, Nick Ryder, Melanie Subbiah, Jared~D Kaplan, Prafulla Dhariwal, Arvind Neelakantan, Pranav Shyam, Girish Sastry, Amanda Askell, Sandhini Agarwal, Ariel Herbert-Voss, Gretchen Krueger, Tom Henighan, Rewon Child, Aditya Ramesh, Daniel Ziegler, Jeffrey Wu, Clemens Winter, Chris Hesse, Mark Chen, Eric Sigler, Mateusz Litwin, Scott Gray, Benjamin Chess, Jack Clark, Christopher Berner, Sam McCandlish, Alec Radford, Ilya Sutskever, and Dario Amodei.
\newblock Language models are few-shot learners.
\newblock In H.~Larochelle, M.~Ranzato, R.~Hadsell, M.F. Balcan, and H.~Lin, editors, {\em Advances in Neural Information Processing Systems}, volume~33, pages 1877--1901. Curran Associates, Inc., 2020.

\bibitem{what_algorithm_is_in_context_learning}
Ekin Aky{\"u}rek, Dale Schuurmans, Jacob Andreas, Tengyu Ma, and Denny Zhou.
\newblock What learning algorithm is in-context learning? investigations with linear models.
\newblock In {\em The Eleventh International Conference on Learning Representations}, 2023.

\bibitem{shallow_arithmetic_arc}
Luca~H. Thoms, Karel~A. Veldkamp, Hannes Rosenbusch, and Claire~E. Stevenson.
\newblock Solving arc visual analogies with neural embeddings and vector arithmetic: A generalized method.
\newblock {\em ArXiv}, abs/2311.08083, 2023.

\bibitem{protonet}
Jake Snell, Kevin Swersky, and Richard~S. Zemel.
\newblock Prototypical networks for few-shot learning.
\newblock {\em CoRR}, abs/1703.05175, 2017.

\bibitem{codeit}
Natasha Butt, Blazej Manczak, Auke Wiggers, Corrado Rainone, David Zhang, Michaël Defferrard, and Taco Cohen.
\newblock Codeit: Self-improving language models with prioritized hindsight replay, 2024.

\bibitem{evidence_of_meaning}
Charles Jin and Martin Rinard.
\newblock Evidence of meaning in language models trained on programs, 2023.

\bibitem{Implicit_Representations_of_Meaning}
Belinda~Z. Li, Maxwell Nye, and Jacob Andreas.
\newblock Implicit representations of meaning in neural language models.
\newblock In Chengqing Zong, Fei Xia, Wenjie Li, and Roberto Navigli, editors, {\em Proceedings of the 59th Annual Meeting of the Association for Computational Linguistics and the 11th International Joint Conference on Natural Language Processing (Volume 1: Long Papers)}, pages 1813--1827, Online, August 2021. Association for Computational Linguistics.

\bibitem{longt5}
Mandy Guo, Joshua Ainslie, David~C. Uthus, Santiago Onta{\~{n}}{\'{o}}n, Jianmo Ni, Yun{-}Hsuan Sung, and Yinfei Yang.
\newblock Longt5: Efficient text-to-text transformer for long sequences.
\newblock {\em CoRR}, abs/2112.07916, 2021.

\bibitem{t5}
Colin Raffel, Noam Shazeer, Adam Roberts, Katherine Lee, Sharan Narang, Michael Matena, Yanqi Zhou, Wei Li, and Peter~J. Liu.
\newblock Exploring the limits of transfer learning with a unified text-to-text transformer.
\newblock {\em CoRR}, abs/1910.10683, 2019.

\bibitem{piaget1952origins_of_intelligence}
Jean Piaget.
\newblock {\em The origins of intelligence in children.}
\newblock The origins of intelligence in children. W W Norton {\&} Co, New York, NY, US, 1952.

\bibitem{bober2024neural}
Mikel Bober-Irizar and Soumya Banerjee.
\newblock Neural networks for abstraction and reasoning: Towards broad generalization in machines.
\newblock {\em arXiv preprint arXiv:2402.03507}, 2024.

\bibitem{general_icl}
Louis Kirsch, James Harrison, Jascha Sohl-Dickstein, and Luke Metz.
\newblock General-purpose in-context learning by meta-learning transformers.
\newblock In {\em Sixth Workshop on Meta-Learning at the Conference on Neural Information Processing Systems}, 2022.

\bibitem{code_which_stage}
Yingwei Ma, Yue Liu, Yue Yu, Yuanliang Zhang, Yu~Jiang, Changjian Wang, and Shanshan Li.
\newblock At which training stage does code data help {LLM}s reasoning?
\newblock In {\em The Twelfth International Conference on Learning Representations}, 2024.

\bibitem{Koepf2022}
A.~Koepf.
\newblock Arc research: Riddle synthesis, 2022.
\newblock Software.

\bibitem{ice_cuber}
Icecuber Top-Quarks.
\newblock Top-quarks/arc-solution: Code for 1st place solution to kaggle’s abstraction and reasoning challenge, 2020.

\bibitem{sun2020testtime}
Yu~Sun, Xiaolong Wang, Zhuang Liu, John Miller, Alexei~A. Efros, and Moritz Hardt.
\newblock Test-time training for out-of-distribution generalization, 2020.

\bibitem{lost_in_the_middle_attention}
Nelson~F. Liu, Kevin Lin, John Hewitt, Ashwin Paranjape, Michele Bevilacqua, Fabio Petroni, and Percy Liang.
\newblock {Lost in the Middle: How Language Models Use Long Contexts}.
\newblock {\em Transactions of the Association for Computational Linguistics}, 12:157--173, 02 2024.

\bibitem{ouyang2022traininglanguagemodelsfollow}
Long Ouyang, Jeff Wu, Xu~Jiang, Diogo Almeida, Carroll~L. Wainwright, Pamela Mishkin, Chong Zhang, Sandhini Agarwal, Katarina Slama, Alex Ray, John Schulman, Jacob Hilton, Fraser Kelton, Luke Miller, Maddie Simens, Amanda Askell, Peter Welinder, Paul Christiano, Jan Leike, and Ryan Lowe.
\newblock Training language models to follow instructions with human feedback, 2022.

\bibitem{flan}
Jason Wei, Maarten Bosma, Vincent Zhao, Kelvin Guu, Adams~Wei Yu, Brian Lester, Nan Du, Andrew~M. Dai, and Quoc~V Le.
\newblock Finetuned language models are zero-shot learners.
\newblock In {\em International Conference on Learning Representations}, 2022.

\bibitem{cot_prompting}
Jason Wei, Xuezhi Wang, Dale Schuurmans, Maarten Bosma, brian ichter, Fei Xia, Ed~H. Chi, Quoc~V Le, and Denny Zhou.
\newblock Chain of thought prompting elicits reasoning in large language models.
\newblock In Alice~H. Oh, Alekh Agarwal, Danielle Belgrave, and Kyunghyun Cho, editors, {\em Advances in Neural Information Processing Systems}, 2022.

\bibitem{gpt_3}
Tom Brown, Benjamin Mann, Nick Ryder, Melanie Subbiah, Jared~D Kaplan, Prafulla Dhariwal, Arvind Neelakantan, Pranav Shyam, Girish Sastry, Amanda Askell, Sandhini Agarwal, Ariel Herbert-Voss, Gretchen Krueger, Tom Henighan, Rewon Child, Aditya Ramesh, Daniel Ziegler, Jeffrey Wu, Clemens Winter, Chris Hesse, Mark Chen, Eric Sigler, Mateusz Litwin, Scott Gray, Benjamin Chess, Jack Clark, Christopher Berner, Sam McCandlish, Alec Radford, Ilya Sutskever, and Dario Amodei.
\newblock Language models are few-shot learners.
\newblock In H.~Larochelle, M.~Ranzato, R.~Hadsell, M.F. Balcan, and H.~Lin, editors, {\em Advances in Neural Information Processing Systems}, volume~33, pages 1877--1901. Curran Associates, Inc., 2020.

\bibitem{lora_first}
Edward~J Hu, yelong shen, Phillip Wallis, Zeyuan Allen-Zhu, Yuanzhi Li, Shean Wang, Lu~Wang, and Weizhu Chen.
\newblock Lo{RA}: Low-rank adaptation of large language models.
\newblock In {\em International Conference on Learning Representations}, 2022.

\bibitem{beam_search}
Clara Meister, Ryan Cotterell, and Tim Vieira.
\newblock If beam search is the answer, what was the question?
\newblock In Bonnie Webber, Trevor Cohn, Yulan He, and Yang Liu, editors, {\em Proceedings of the 2020 Conference on Empirical Methods in Natural Language Processing (EMNLP)}, pages 2173--2185, Online, November 2020. Association for Computational Linguistics.

\bibitem{temperature_effect}
Matthew Renze.
\newblock The effect of sampling temperature on problem solving in large language models.
\newblock In Yaser Al-Onaizan, Mohit Bansal, and Yun-Nung Chen, editors, {\em Findings of the Association for Computational Linguistics: EMNLP 2024}, pages 7346--7356, Miami, Florida, USA, November 2024. Association for Computational Linguistics.

\bibitem{alpha_code}
Yujia Li, David Choi, Junyoung Chung, Nate Kushman, Julian Schrittwieser, Rémi Leblond, Tom Eccles, James Keeling, Felix Gimeno, Agustin~Dal Lago, Thomas Hubert, Peter Choy, Cyprien de~Masson~d’Autume, Igor Babuschkin, Xinyun Chen, Po-Sen Huang, Johannes Welbl, Sven Gowal, Alexey Cherepanov, James Molloy, Daniel~J. Mankowitz, Esme~Sutherland Robson, Pushmeet Kohli, Nando de~Freitas, Koray Kavukcuoglu, and Oriol Vinyals.
\newblock Competition-level code generation with alphacode.
\newblock {\em Science}, 378(6624):1092--1097, 2022.

\bibitem{arc_agi_2024}
Francois Chollet, Mike Knoop, Bryan Landers, Greg Kamradt, Hansueli Jud, Walter Reade, and Addison Howard.
\newblock Arc prize 2024.
\newblock \url{https://kaggle.com/competitions/arc-prize-2024}, 2024.
\newblock Kaggle.

\bibitem{Knoop_2024}
Mike Knoop.
\newblock Introducing the arc-agi public leaderboard, Jun 2024.

\bibitem{arc_agi_website}
Francois Chollet, Mike Knoop, Bryan Landers, Greg Kamradt, Hansueli Jud, Walter Reade, and Addison Howard.
\newblock Arc prize website 2025.
\newblock \url{https://arcprize.org}, 2025.
\newblock ARC-AGI.

\bibitem{arcathon2023}
Arcathon winners 2023.
\newblock \url{https://lab42.global/winners/}, 2023.
\newblock Accessed: 2025-03-05.

\bibitem{Kaplan2020ScalingLF}
Jared Kaplan, Sam McCandlish, Tom Henighan, Tom~B. Brown, Benjamin Chess, Rewon Child, Scott Gray, Alec Radford, Jeff Wu, and Dario Amodei.
\newblock Scaling laws for neural language models.
\newblock {\em ArXiv}, abs/2001.08361, 2020.

\bibitem{chain_of_thought_prompting}
Jason Wei, Xuezhi Wang, Dale Schuurmans, Maarten Bosma, Brian Ichter, Fei Xia, Ed~H. Chi, Quoc~V. Le, and Denny Zhou.
\newblock Chain-of-thought prompting elicits reasoning in large language models.
\newblock In {\em Proceedings of the 36th International Conference on Neural Information Processing Systems}, NIPS '22, Red Hook, NY, USA, 2024. Curran Associates Inc.

\bibitem{hermann2023foundations}
Katherine~L Hermann, Hossein Mobahi, Thomas Fel, and Michael~C Mozer.
\newblock On the foundations of shortcut learning.
\newblock {\em arXiv preprint arXiv:2310.16228}, 2023.

\bibitem{shah2020pitfalls}
Harshay Shah, Kaustav Tamuly, Aditi Raghunathan, Prateek Jain, and Praneeth Netrapalli.
\newblock The pitfalls of simplicity bias in neural networks.
\newblock {\em Advances in Neural Information Processing Systems}, 33:9573--9585, 2020.

\bibitem{sedementation_paper}
Thomas Fel, Louis Bethune, Andrew~Kyle Lampinen, Thomas Serre, and Katherine Hermann.
\newblock Understanding visual feature reliance through the lens of complexity, 2024.

\bibitem{data_distributional_properties_ICL}
Stephanie~C.Y. Chan, Adam Santoro, Andrew~Kyle Lampinen, Jane~X Wang, Aaditya~K Singh, Pierre~Harvey Richemond, James McClelland, and Felix Hill.
\newblock Data distributional properties drive emergent in-context learning in transformers.
\newblock In Alice~H. Oh, Alekh Agarwal, Danielle Belgrave, and Kyunghyun Cho, editors, {\em Advances in Neural Information Processing Systems}, 2022.

\bibitem{Lake2023Humanlike}
B.M. Lake and M.~Baroni.
\newblock Human-like systematic generalization through a meta-learning neural network.
\newblock {\em Nature}, 623:115--121, 2023.

\bibitem{the_surp}
Ekin Akyürek, Mehul Damani, Linlu Qiu, Han Guo, Yoon Kim, and Jacob Andreas.
\newblock The surprising effectiveness of test-time training for abstract reasoning, 2024.

\bibitem{community_post}
Jack Cole and Mohamed Osman.
\newblock Dataset-induced meta-learning (and other tricks): Improving model efficiency on arc.
\newblock \url{https://lab42.global/community-model-efficiency/}, 2023.
\newblock Accessed: March 3, 2025.

\bibitem{mlst_ep}
Jack Cole, Mohamed Osman, Michael Hodel, Keith Duggar, and Tim Scarfe.
\newblock Machine learning street talk.
\newblock \url{https://www.youtube.com/watch?v=jSAT_RuJ_Cg}, 2024.
\newblock Accessed: June 2024.

\bibitem{transduction_induction}
Wen-Ding Li, Keya Hu, Carter Larsen, Yuqing Wu, Simon Alford, Caleb Woo, Spencer~M. Dunn, Hao Tang, Michelangelo Naim, Dat Nguyen, Wei-Long Zheng, Zenna Tavares, Yewen Pu, and Kevin Ellis.
\newblock Combining induction and transduction for abstract reasoning, 2024.

\bibitem{palm}
Aakanksha Chowdhery, Sharan Narang, Jacob Devlin, Maarten Bosma, Gaurav Mishra, Adam Roberts, Paul Barham, Hyung~Won Chung, Charles Sutton, Sebastian Gehrmann, Parker Schuh, Kensen Shi, Sasha Tsvyashchenko, Joshua Maynez, Abhishek Rao, Parker Barnes, Yi~Tay, Noam Shazeer, Vinodkumar Prabhakaran, Emily Reif, Nan Du, Ben Hutchinson, Reiner Pope, James Bradbury, Jacob Austin, Michael Isard, Guy Gur-Ari, Pengcheng Yin, Toju Duke, Anselm Levskaya, Sanjay Ghemawat, Sunipa Dev, Henryk Michalewski, Xavier Garcia, Vedant Misra, Kevin Robinson, Liam Fedus, Denny Zhou, Daphne Ippolito, David Luan, Hyeontaek Lim, Barret Zoph, Alexander Spiridonov, Ryan Sepassi, David Dohan, Shivani Agrawal, Mark Omernick, Andrew~M. Dai, Thanumalayan~Sankaranarayana Pillai, Marie Pellat, Aitor Lewkowycz, Erica Moreira, Rewon Child, Oleksandr Polozov, Katherine Lee, Zongwei Zhou, Xuezhi Wang, Brennan Saeta, Mark Diaz, Orhan Firat, Michele Catasta, Jason Wei, Kathy Meier-Hellstern, Douglas Eck, Jeff Dean, Slav Petrov, and Noah Fiedel.
\newblock Palm: Scaling language modeling with pathways.
\newblock {\em Journal of Machine Learning Research}, 24(240):1--113, 2023.

\bibitem{touvron2023llama}
Hugo Touvron, Thibaut Lavril, Gautier Izacard, Xavier Martinet, Marie-Anne Lachaux, Timothée Lacroix, Baptiste Rozière, Naman Goyal, Eric Hambro, Faisal Azhar, Aurelien Rodriguez, Armand Joulin, Edouard Grave, and Guillaume Lample.
\newblock Llama: Open and efficient foundation language models, 2023.

\bibitem{HELM_full_eval}
Percy Liang, Rishi Bommasani, Tony Lee, Dimitris Tsipras, Dilara Soylu, Michihiro Yasunaga, Yian Zhang, Deepak Narayanan, Yuhuai Wu, Ananya Kumar, Benjamin Newman, Binhang Yuan, Bobby Yan, Ce~Zhang, Christian~Alexander Cosgrove, Christopher~D Manning, Christopher Re, Diana Acosta-Navas, Drew~Arad Hudson, Eric Zelikman, Esin Durmus, Faisal Ladhak, Frieda Rong, Hongyu Ren, Huaxiu Yao, Jue WANG, Keshav Santhanam, Laurel Orr, Lucia Zheng, Mert Yuksekgonul, Mirac Suzgun, Nathan Kim, Neel Guha, Niladri~S. Chatterji, Omar Khattab, Peter Henderson, Qian Huang, Ryan~Andrew Chi, Sang~Michael Xie, Shibani Santurkar, Surya Ganguli, Tatsunori Hashimoto, Thomas Icard, Tianyi Zhang, Vishrav Chaudhary, William Wang, Xuechen Li, Yifan Mai, Yuhui Zhang, and Yuta Koreeda.
\newblock Holistic evaluation of language models.
\newblock {\em Transactions on Machine Learning Research}, 2023.
\newblock Featured Certification, Expert Certification.

\bibitem{code_commonsense_few_shot}
Aman Madaan, Shuyan Zhou, Uri Alon, Yiming Yang, and Graham Neubig.
\newblock Language models of code are few-shot commonsense learners.
\newblock In Yoav Goldberg, Zornitsa Kozareva, and Yue Zhang, editors, {\em Proceedings of the 2022 Conference on Empirical Methods in Natural Language Processing}, pages 1384--1403, Abu Dhabi, United Arab Emirates, December 2022. Association for Computational Linguistics.

\bibitem{code_ablate}
Anonymous.
\newblock How does code pretraining affect language model task performance?
\newblock {\em Submitted to Transactions on Machine Learning Research}, 2024.
\newblock Under review.

\bibitem{mirchandani2023large}
Suvir Mirchandani, Fei Xia, Pete Florence, brian ichter, Danny Driess, Montserrat~Gonzalez Arenas, Kanishka Rao, Dorsa Sadigh, and Andy Zeng.
\newblock Large language models as general pattern machines.
\newblock In {\em 7th Annual Conference on Robot Learning}, 2023.

\bibitem{mitchell2023comparing}
Melanie Mitchell, Alessandro~B. Palmarini, and Arsenii~Kirillovich Moskvichev.
\newblock Comparing humans, {GPT}-4, and {GPT}-4v on abstraction and reasoning tasks.
\newblock In {\em AAAI 2024 Workshop on ''Are Large Language Models Simply Causal Parrots?''}, 2023.

\bibitem{moskvichev2023the}
Arsenii~Kirillovich Moskvichev, Victor~Vikram Odouard, and Melanie Mitchell.
\newblock The concept{ARC} benchmark: Evaluating understanding and generalization in the {ARC} domain.
\newblock {\em Transactions on Machine Learning Research}, 2023.

\bibitem{10208934}
G.~Camposampiero, L.~Houmard, B.~Estermann, J.~Mathys, and R.~Wattenhofer.
\newblock Abstract visual reasoning enabled by language.
\newblock In {\em 2023 IEEE/CVF Conference on Computer Vision and Pattern Recognition Workshops (CVPRW)}, pages 2643--2647, Los Alamitos, CA, USA, jun 2023. IEEE Computer Society.

\bibitem{xu2024llms}
Yudong Xu, Wenhao Li, Pashootan Vaezipoor, Scott Sanner, and Elias~Boutros Khalil.
\newblock {LLM}s and the abstraction and reasoning corpus: Successes, failures, and the importance of object-based representations.
\newblock {\em Transactions on Machine Learning Research}, 2024.

\bibitem{wang2024hypothesis}
Ruocheng Wang, Eric Zelikman, Gabriel Poesia, Yewen Pu, Nick Haber, and Noah Goodman.
\newblock Hypothesis search: Inductive reasoning with language models.
\newblock In {\em The Twelfth International Conference on Learning Representations}, 2024.

\bibitem{butt2024codeit}
Natasha Butt, Blazej Manczak, Auke Wiggers, Corrado Rainone, David Zhang, Michaël Defferrard, and Taco Cohen.
\newblock Codeit: Self-improving language models with prioritized hindsight replay, 2024.

\bibitem{lei2024generalized}
Chao Lei, Nir Lipovetzky, and Krista~A. Ehinger.
\newblock Generalized planning for the abstraction and reasoning corpus, 2024.

\bibitem{dreamcoder}
Kevin Ellis, Catherine Wong, Maxwell Nye, Mathias Sabl\'{e}-Meyer, Lucas Morales, Luke Hewitt, Luc Cary, Armando Solar-Lezama, and Joshua~B. Tenenbaum.
\newblock Dreamcoder: bootstrapping inductive program synthesis with wake-sleep library learning.
\newblock In {\em Proceedings of the 42nd ACM SIGPLAN International Conference on Programming Language Design and Implementation}, PLDI 2021, page 835–850, New York, NY, USA, 2021. Association for Computing Machinery.

\bibitem{assouel2022objectcentric}
Rim Assouel, Pau Rodriguez, Perouz Taslakian, David Vazquez, and Yoshua Bengio.
\newblock Object-centric compositional imagination for visual abstract reasoning.
\newblock In {\em ICLR2022 Workshop on the Elements of Reasoning: Objects, Structure and Causality}, 2022.

\bibitem{Ouellette2023}
S.~Ouellette.
\newblock Arc\_gym: A data generation framework for the abstraction \& reasoning corpus, 2023.
\newblock Software.

\bibitem{Hodel2024}
M.~Hodel.
\newblock Addressing the abstraction and reasoning corpus via procedural example generation, 2024.
\newblock Preprint.

\bibitem{Hupkes2020}
D.~Hupkes, V.~Dankers, M.~Mul, and E.~Bruni.
\newblock Compositionality decomposed: How do neural networks generalise?
\newblock {\em Journal of Artificial Intelligence Research}, 67:757--795, 2020.

\end{thebibliography}

\newpage

\appendix

\section{Training Dataset Construction and Composition}

Our approach to solving the Abstraction and Reasoning Corpus (ARC) Challenge incorporated a
diverse set of training datasets,
combining both existing public resources and custom-generated synthetic data.
This data strategy was designed to develop robust abstract reasoning capabilities and improve
grid-based pattern recognition tasks.

\paragraph{Multimodal Grid Translation Datasets:} These datasets facilitate translation between visual
and symbolic representations of grids - including Base64 images, text, English descriptions,
and code implementations. They focus on simple shapes and positions to bridge visual and symbolic
reasoning relevant to ARC.

\paragraph{Extended PCFG Datasets:} Building upon the principles of Probabilistic Context-Free Grammars
(PCFGs), we use an expanded PCFG dataset with 100 distinct string operations, designed to resemble
ARC riddles. These datasets incorporate program synthesis components, emphasizing function
name generation as part of the learning process.

\paragraph{Cellular Automata and Mathematical Pattern Datasets:} We generate cellular automata tasks within
the ARC framework and create riddle boards using mathematical equations. These datasets introduce
varied pattern complexities and cross-item riddles that require learning underlying concepts
across multiple examples. Repair-type riddles are also generated to induce priors related to
fundamental visual reasoning concepts like symmetries, shapes, progression, and counting.

Our data generation strategy also leverages frameworks such as ARC\_gym \cite{Ouellette2023} for supplementary
examples with customizable grid characteristics. We also enhance existing datasets, including
augmented ConceptARC data \cite{concept_arc_benchmark} and augmented official public ARC datasets.
The recent integration of RE-ARC data \cite{Hodel2024}, with its procedurally generated examples
for the 400 ARC training tasks, has further improved performance by increasing exposure to
task-specific patterns and transformations.

\subsection{Public Datasets}

The training pipeline incorporated several established datasets from the research community:

\subsubsection{Language and Reasoning Datasets}
\begin{enumerate}
\item WizardLM Evolution Instruct V2 (196k examples)
\item TinyStories-GPT4 (150,000 examples)
\item SQuAD v2
\item Chain-of-Thought Submix
\item Open-Platypus
\item SlimOrca
\end{enumerate}

\subsubsection{Domain-Specific Scientific Datasets}
\begin{enumerate}
\item Arxiv Math Instruct (50k examples)
\item Arxiv CS/ML Instruct (50k examples)
\item MathInstruct
\item Arxiv Physics Instruct (30k examples)
\end{enumerate}

\subsubsection{Multimodal and Visual Reasoning Datasets}
We integrated several multimodal reasoning datasets from the M3IT collection, specifically focusing on:
\begin{enumerate}
\item CLEVR
\item NLVR
\item VCR
\item Visual-MRC
\end{enumerate}

\subsubsection{Programming and Code-Related Datasets}
\begin{enumerate}
\item Magicoder Evolution Instruct (110K examples)
\item Open Instruct v1 (derived from OASST and Dolly HHRLHF)
\end{enumerate}

\begin{figure*}[h]  %
  \centering
  \subcaptionbox{Mirror-removal\label{fig:mirror}}
    [.30\linewidth]{\includegraphics[width=\linewidth]{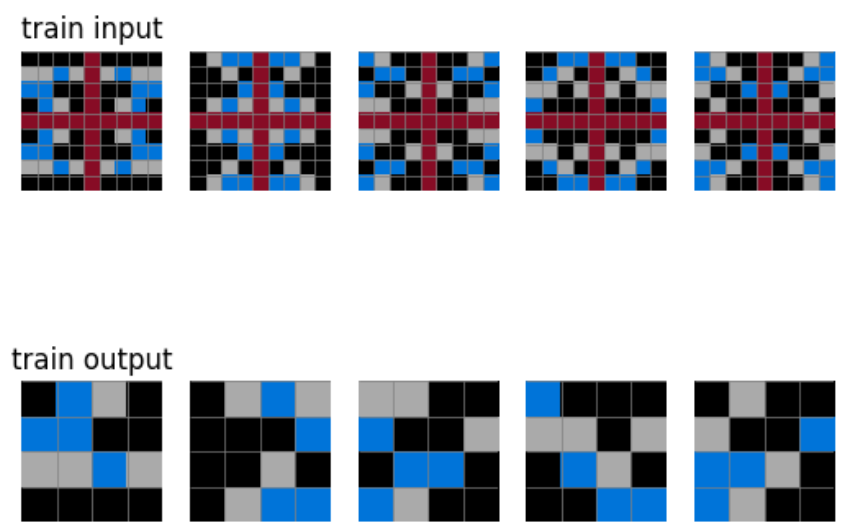}}\hfill
  \subcaptionbox{Fill-in-the-shapes\label{fig:fillshapes}}
    [.30\linewidth]{\includegraphics[width=\linewidth]{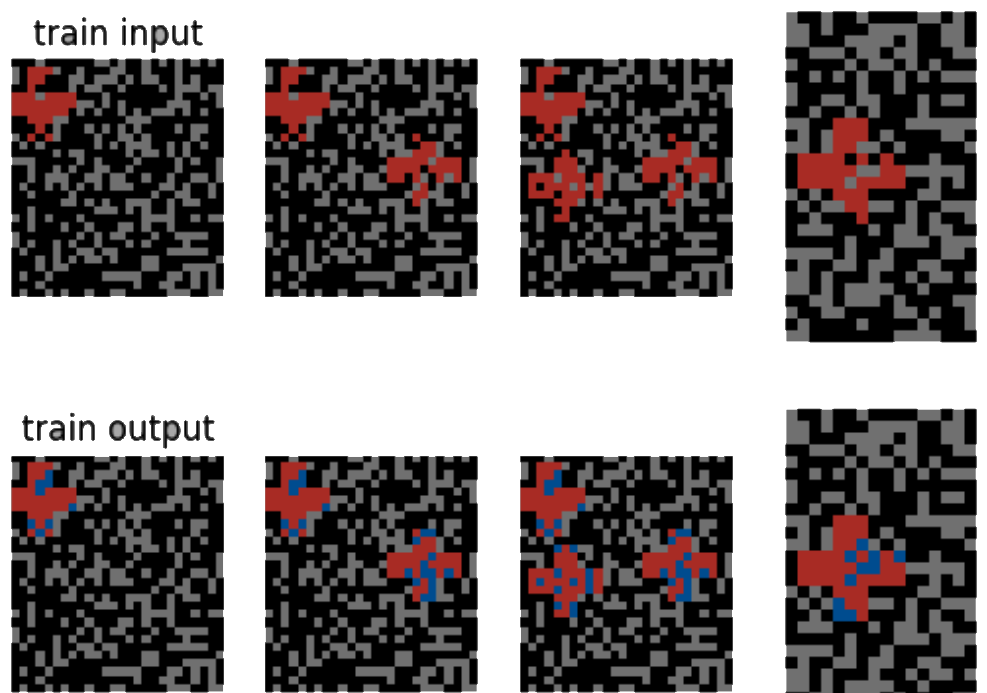}}\hfill
  \subcaptionbox{Fractal (multi-colour)\label{fig:fractalcolor}}
    [.30\linewidth]{\includegraphics[width=\linewidth]{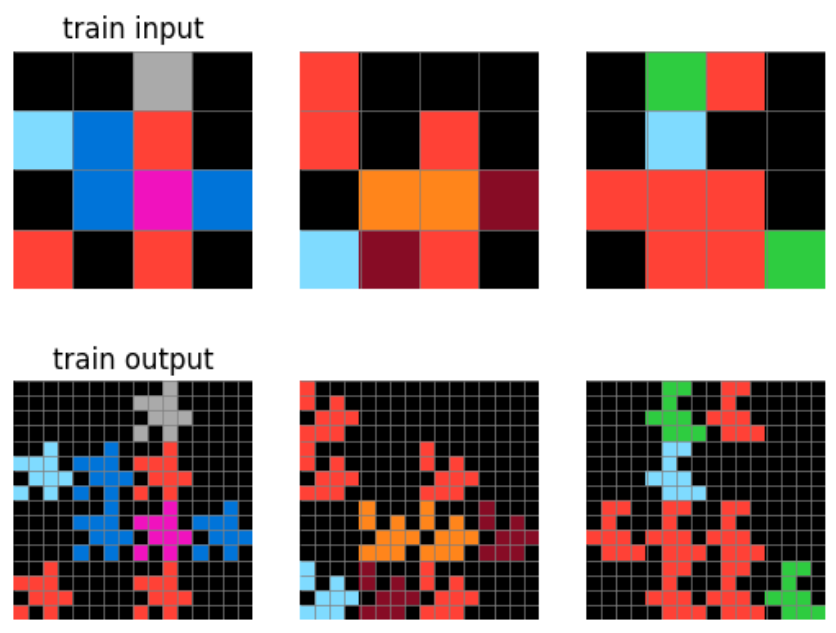}}

  \vspace{1em}

  \subcaptionbox{Fractal (mono)\label{fig:fractalm}}
    [.30\linewidth]{\includegraphics[width=\linewidth]{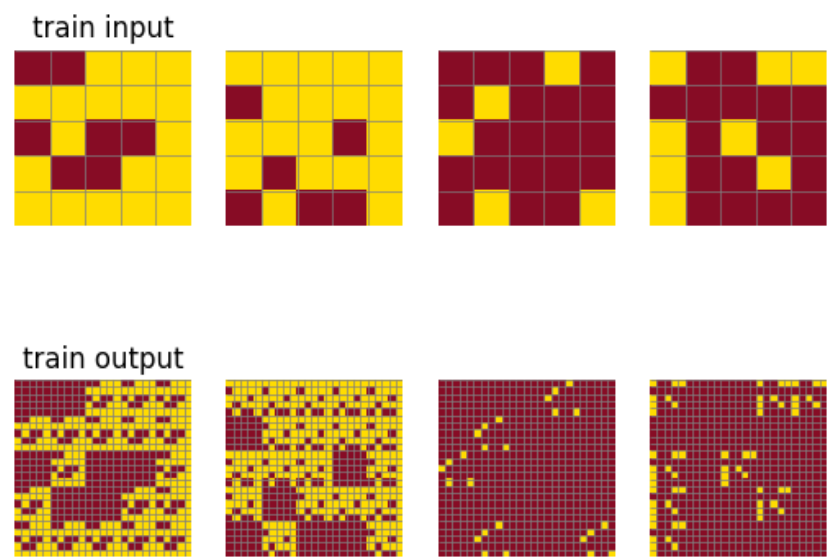}}\hfill
  \subcaptionbox{Core-concept rules\label{fig:coreconcept}}
    [.30\linewidth]{\includegraphics[width=\linewidth]{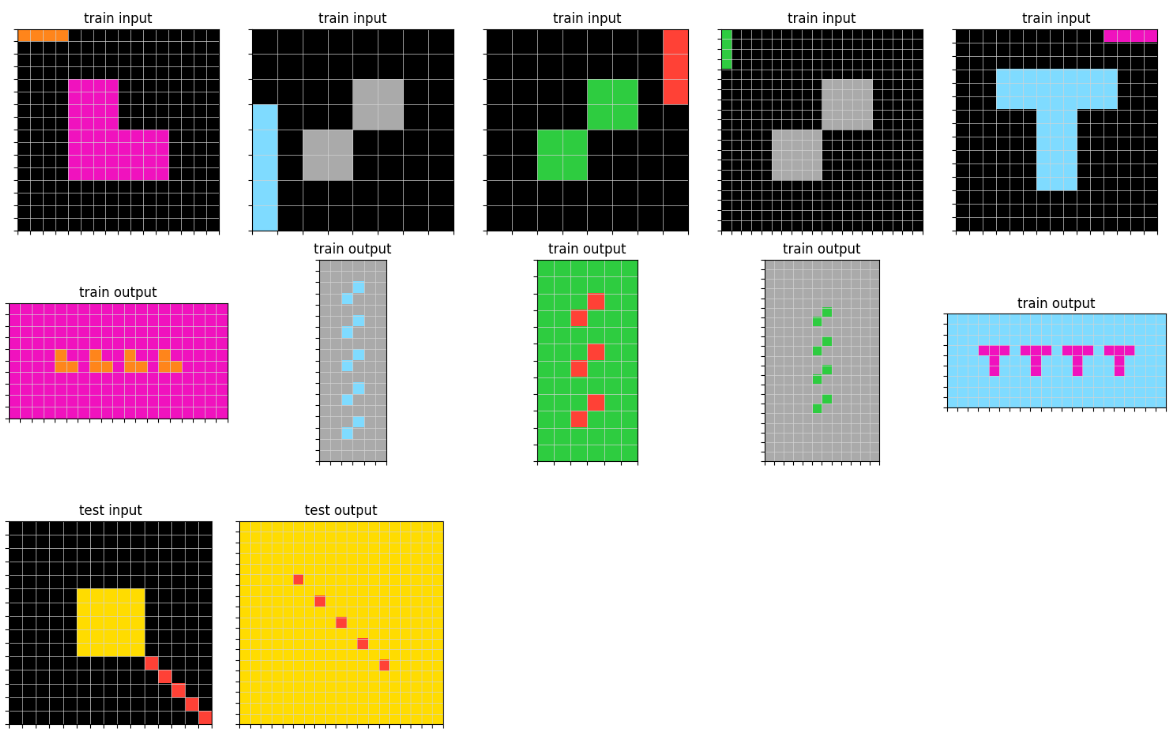}}\hfill
  \subcaptionbox{Area-repair (dense)\label{fig:areadense}}
    [.30\linewidth]{\includegraphics[width=\linewidth]{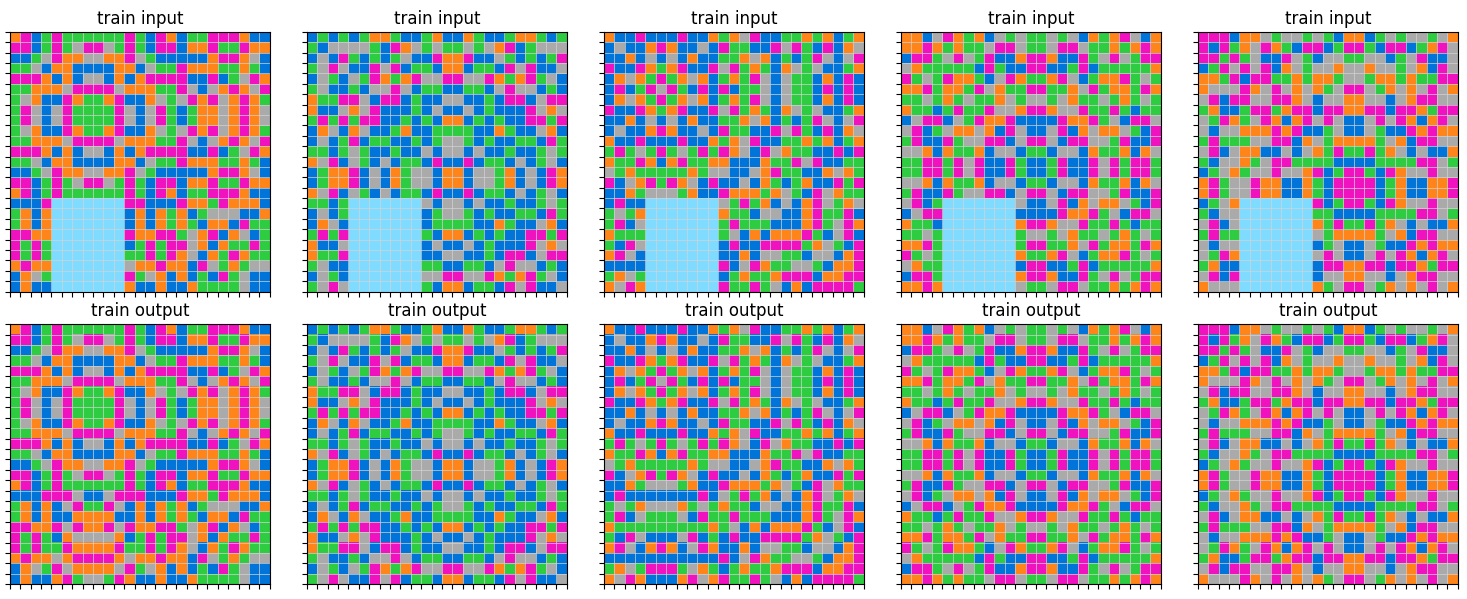}}

  \vspace{1em}

  \subcaptionbox{Area-repair (sparse)\label{fig:areasparse}}
    [.30\linewidth]{\includegraphics[width=\linewidth]{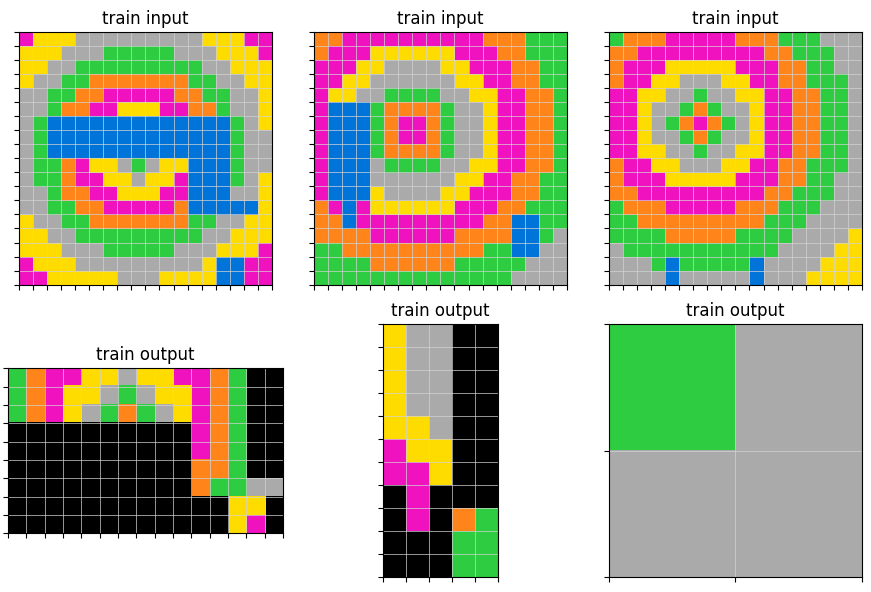}}\hfill
  \subcaptionbox{Cellular-automata\label{fig:cellauto}}
    [.30\linewidth]{\includegraphics[width=\linewidth]{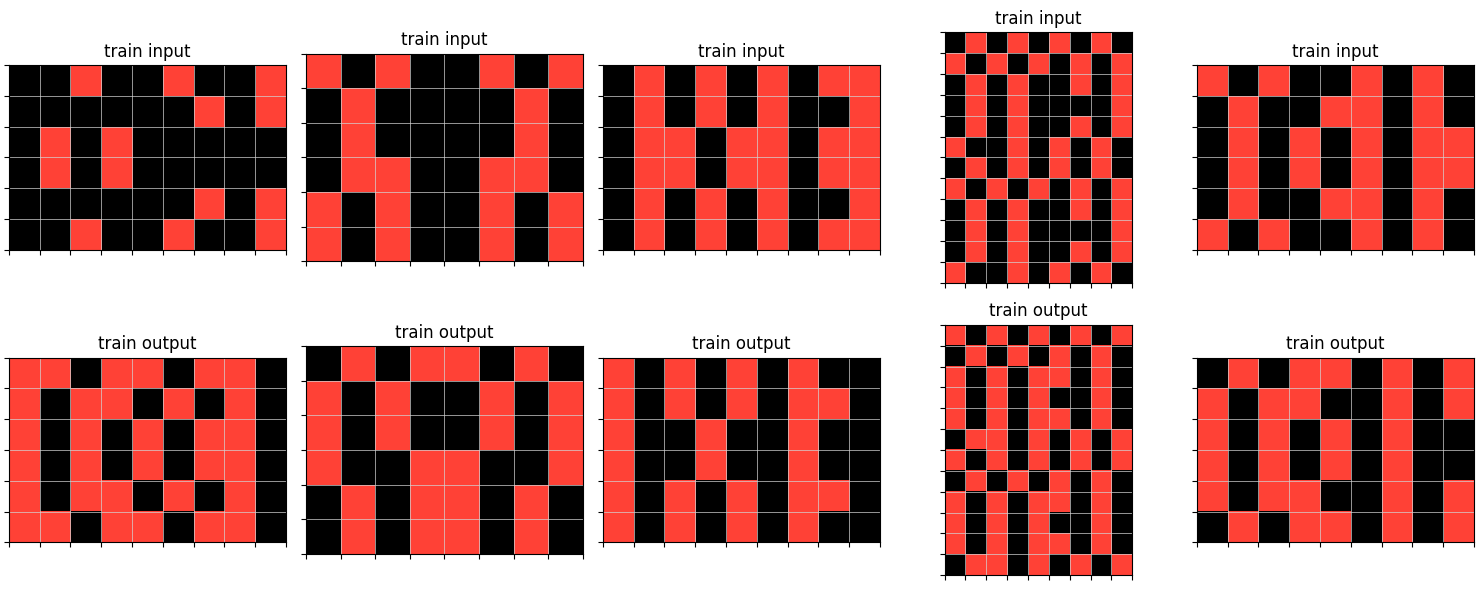}}

  \caption{Gallery of synthetic ARC-style riddles used in pre-training:
           symmetry repair, noise-robust completion, fractal growth,
         core-concept generalisation, area repair, and cellular-automata evolution with simple rules.}
  \label{fig:synthetic-gallery}
\end{figure*}

\subsection{Custom Synthetic Datasets}

\textbf{1. Arithmetic and Counting Tasks}

We created fine-grained arithmetic and counting datasets that emphasized precise numerical operations and pattern recognition. These datasets were designed to strengthen the model's capability in handling discrete mathematical operations commonly found in ARC tasks.

\textbf{2. Multimodal Grid Translation}

We developed a specialized dataset for translating between:
\begin{enumerate}
\item Base64 images of grids
\item Text representations
\item English descriptions
\item Code implementations
\end{enumerate}

This dataset focused on simple shapes and their precise positions, creating a bridge between visual and symbolic reasoning.

\textbf{3. Extended PCFG Dataset}

Following work on neural networks' compositional capabilities \cite{Hupkes2020}
and the demonstrated effectiveness of Probabilistic Context-Free Grammars (PCFGs)
for pattern learning \cite{general_pattern_machines}, we significantly expanded our PCFG dataset to include:
\begin{enumerate}
\item 100 distinct string operations with few-shot examples
\item Tasks structured similarly to ARC riddles
\item Program synthesis components requiring function name generation
\end{enumerate}

These tasks were generally designed to be more challenging than typical ARC problems, pushing the boundaries of abstract reasoning capabilities.

\textbf{4. Multi-Task Integration Dataset}

We created a synthetic multi-task dataset that combined multiple problem types within single training examples:
\begin{enumerate}
\item Arithmetic code generation
\item Chained boolean expression evaluation
\item PCFG tasks
\end{enumerate}

The dataset was structured in JSON format, with both prompts and answers carefully formatted for consistency.

\textbf{5. Cellular Automata and Mathematical Patterns}

We incorporated additional pattern-generation approaches:
\begin{enumerate}
\item Cellular automata tasks within the ARC framework
\item ARC riddle boards generated using mathematical equations
\item Pattern complexity variations to induce diverse priors
\item Cross-item riddles requiring learning underlying concepts across multiple examples
\item Repair-type riddles for inducing priors including symmetries, shapes, progression, and counting
\end{enumerate}

See \autoref{fig:synthetic-gallery} for examples of synthetic ARC-style riddles used in pre-training.

\subsection{Data Sources and Generation}

We incorporated several data sources and generation approaches:

\begin{enumerate}
\item \textbf{Framework-Generated Data:}
\begin{enumerate}
  \item ARC\_gym framework \cite{Ouellette2023} to generate supplementary training examples with:
  \begin{enumerate}
  \item Customizable grid sizes and complexity levels
  \item Controlled out-of-distribution task generation
  \item Variable numbers of primitives per task
  \item Systematic composition of basic operations
  \end{enumerate}
\item IceCuber-based riddle generation \cite{Koepf2022}, including:
  \begin{enumerate}
  \item Synthetic riddle generation
  \item Solution graph prediction tasks
  \end{enumerate}
\end{enumerate}

\item \textbf{Enhanced Public Datasets:}
\begin{enumerate}
  \item Augmented ConceptARC data \cite{concept_arc_benchmark}, leveraging:
  \begin{enumerate}
  \item Concept group organization
  \item Varying complexity levels
  \item Abstraction-focused variations
  \end{enumerate}
\item Augmented official public ARC datasets
\item Modified versions of publicly available ARC\_gym generated data
\end{enumerate}
\end{enumerate}

\subsection{Recent Additions}

In early 2024, we integrated data from RE-ARC \cite{Hodel2024}, which provided procedurally generated examples for the 400 ARC training tasks. This addition contributed to improved performance through increased exposure to task-specific patterns and transformations.

\subsection{Training System Notes}

The training system utilizes prepared datasets in chunks for roughly 500,000 training examples.
The chunks are needed to reduce CPU memory usage for processing. The order of the training examples is shuffled.
It automatically incorporates new datasets without interrupting training and some datasets include instruction-following and chat-based interaction capabilities.

\end{document}